\documentclass[10pt,twocolumn,letterpaper]{article}

\usepackage[pagenumbers]{cvpr} % To force page numbers, e.g. for an arXiv version
\usepackage{algorithm}
\usepackage{algorithmic}

\definecolor{cvprblue}{rgb}{0.21,0.49,0.74}
\usepackage[pagebackref,breaklinks,colorlinks,allcolors=cvprblue]{hyperref}

\def\paperID{1796} % *** Enter the Paper ID here
\def\confName{CVPR}
\def\confYear{2026}

\title{Reversible Efficient Diffusion for Image Fusion}

\author{\textbf{Xingxin Xu}$^{1}$, \textbf{Bing Cao}$^2$, \textbf{DongDong Li}$^5$, \textbf{Qinghua Hu}$^{2}$, \textbf{Pengfei Zhu}$^{2,3,4}$ \\ \\
\textsuperscript{\rm 1}School of New Media and Communication, Tianjin University\\
 \textsuperscript{\rm 2} School of Artificial Intelligence, Tianjin University\\
  \textsuperscript{\rm 3} Low-Altitude Intelligence Laboratory, Xiong'an National Innovation Center  \\
   \textsuperscript{\rm 4} Xiong'an Guochuang Lantian Technology Co., Ltd. \\
 \textsuperscript{\rm 5}  National University of Defense Technology\\
{\tt\small    \{xuxingxin, caobing, huqinghua, zhupengfei\}@tju.edu.cn, lidongdong12@nudt.edu.cn }
}

\begin{document}
\maketitle
\begin{abstract}

Multi-modal image fusion aims to consolidate complementary information from diverse source images into a unified representation. The fused image is expected to preserve fine details and maintain high visual fidelity.
While diffusion models have demonstrated impressive generative capabilities in image generation, they often suffer from detail loss when applied to image fusion tasks. This issue arises from the accumulation of noise errors inherent in the Markov process, leading to inconsistency and degradation in the fused results.
However, incorporating explicit supervision into end-to-end training of diffusion-based image fusion introduces challenges related to computational efficiency.
To address these limitations, we propose the \textit{Reversible Efficient Diffusion} (\textbf{RED}) model —an explicitly supervised training framework that inherits the powerful generative capability of diffusion models while avoiding the distribution estimation. 
 % Existing methods primarily adopt end-to-end training frameworks but typically rely on a single-pass fusion process, which inadequately preserves complementary multi-modal characteristics and often results in suboptimal performance. 
 % RED can dynamically update fusion outputs at each iteration by incorporating newly synthesized feature representations, thereby enabling continuous quality improvement. 
First, we introduce a reversible fusion paradigm within the diffusion process to substantially reduce memory usage, enabling end-to-end training for direct image fusion. Additionally, we integrate a reversible residual block to further reduce memory consumption during training, allowing the model to generate high-quality fusion results. Extensive experimental results demonstrate that RED significantly outperforms state-of-the-art methods across a variety of multi-modal image fusion tasks.

\end{abstract}     
\section{Introduction}
\label{sec:intro}
% \begin{figure*}
%     \centering
%     \vspace{-10pt}
%     \includegraphics[width=1\textwidth]{images/fig_fig_motivation_comp.pdf}  % 调整图片大小
%     \vspace{-5pt}
%     \caption{Illustration comparing the proposed RED framework with previous fusion paradigms. The area of each bubble represents the memory usage of the corresponding model. RED demonstrates superior performance while maintaining both low inference time and low memory usage.}
%     \label{fig:motiv}
% \end{figure*}
\begin{figure}
    \centering
    % \vspace{-10pt}
    \includegraphics[width=0.48\textwidth]{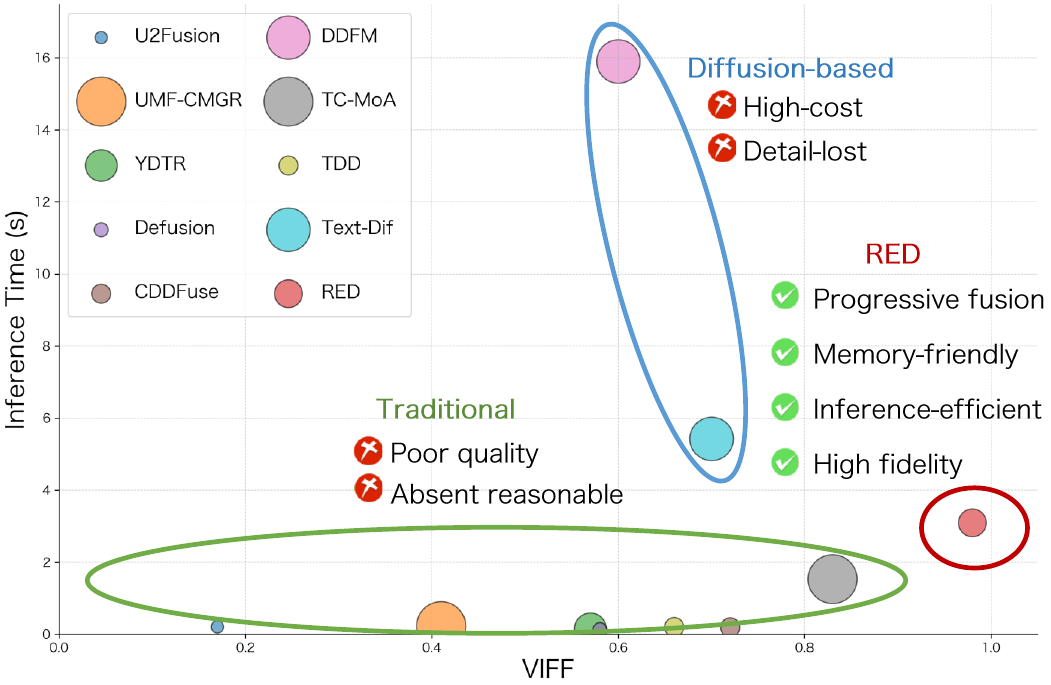}  % 调整图片大小
    \vspace{-15pt}
    \caption{Illustration comparing the proposed RED framework with previous fusion paradigms. The area of each bubble represents the memory usage of the corresponding model. RED demonstrates superior performance while maintaining both low inference time and low memory usage.}
    \label{fig:motiv}
    \vspace{-20pt}
\end{figure}

Multi-modal image fusion is a widely used technique that aims to generate a comprehensive and informative representation by integrating complementary information from diverse source modalities. Representative examples include visible-infrared image fusion (VIF) and medical image fusion (MIF). Its primary objective is to produce fused images that preserve and enhance the most salient features of each modality while eliminating redundant or conflicting information. Multi-modal image fusion has been extensively applied in downstream computer vision tasks, including object detection, semantic segmentation, and disease diagnosis, where it provides valuable insights by combining rich information from multiple sensors. 
% 图像生成领域中表现好，但是在图像融合中表现一般（效率 性能）
% 造成这个现象的原因是由于扩散模型对于隐式（间接） ，缺乏对原始图像的监督
% 

Recent research has explored various fusion techniques, including encoder-decoder fusion, transformer-based fusion, generative adversarial networks (GAN-based) fusion, and more recently, diffusion-based fusion. 
% Although most methods have achieved superior fusion performance, most conduct fusion or reconstruction only once, which may limit their ability to fully capture relevant feature information, potentially resulting in suboptimal outcomes. 
Among generative models, denoising diffusion models (DMs) have attracted wide attention for modeling complex data via a neural-network–parameterized reverse Markov chain~\cite{ho2020denoising}. Latent variants such as Stable Diffusion have achieved impressive image synthesis results~\cite{rombach2022high, ruiz2023dreambooth}.
However, image fusion differs from free-form generation and demands faithful preservation of structural details and strict cross-modal consistency. Vanilla DMs are trained with a noise-prediction objective and typically rely on stochastic sampling, and without explicit structural or semantic constraints tied to the source modalities the fused outputs can lose fine details or exhibitin consistencies~\cite{wang2023diffusion}. Moreover, the multi-step reverse process can accumulate prediction errors across iterations, especially when guidance is weak, introducing artifacts in the final image. These factors limit the direct applicability of off-the-shelf diffusion models to high-fidelity fusion, where explicit structure preservation and modality alignment are critical.
Therefore, achieving effective fusion with diffusion models typically calls for explicit loss constraints and reduced sampling stochasticity, rather than relying solely on standard noise-prediction training.
% While introducing explicit constraints between the input and output may potentially improve application in image fusion with diffusion models, it often incurs significant memory overhead, especially during training. This makes such approaches impractical in real-world scenarios, particularly when working with long diffusion chains.

To address the aforementioned challenges, we propose a novel Reversible Efficient Diffusion (RED) for Image Fusion. RED bridges generative diffusion and fusion objectives by enabling end-to-end training under strong, task-aware supervision from both source images and the fused target, resulting in superior performance, as illustrated in Fig.~\ref{fig:motiv}.
% It effectively overcomes the limitations of existing diffusion models, particularly in implicit noise estimation learning. 
However, naïvely training diffusion models end to end is often prohibitive on consumer-grade hardware due to memory and compute demands. Inspired by invertible architectures~\cite{behrmann2019invertible, zhao2023re2tal}, we propose the reversible fusion paradigm that supervises the entire diffusion process with manageable memory. This is achieved by avoiding the storage of intermediate features during forward propagation and instead recomputing them during backpropagation. 
Furthermore, we incorporate reversible residual blocks~\cite{gomez2017reversible} into the diffusion backbone, providing additional memory savings while maintaining, and even enhancing, the quality of the fused results. 
This design makes RED compatible with explicit fusion losses and reduced-stochasticity sampling, aligning the training objective with fusion-specific requirements without incurring prohibitive memory overhead improved the fidelity and modality consistency of the fused results.
% This capability allows for more accurate image reconstruction, preserving rich and complete information essential for high-quality fusion. 
% Additionally, we incorporate a Prior-Guidance Module, which introduces supplementary prior information to simplify the diffusion process and improve the quality of the fused output.
In summary, the main contributions are as follows:
\begin{itemize}
    \item We propose a reversible efficient diffusion (\textbf{RED}), a novel non-Markovian diffusion training paradigm tailored for image fusion, which leverages supervision from the source images to alleviate noise accumulation and improve structural consistency.
    \item We introduce a reversible fusion strategy that enables memory-efficient training of RED with alternating iterative fusion. In addition, we incorporate reversible residual blocks to further reduce memory overhead and enhance the quality of the fused outputs.
    
 % by avoiding the storage of intermediate activations
    % (基于这个思想，我们提出了---instead 其他缺点）（解释为什么要怎么做）
    \item Extensive quantitative and qualitative experiments demonstrate the effectiveness and generalizability of RED across both visible-infrared and medical image fusion tasks, while also facilitating improvements in downstream applications.
     % \item We propose RED, a novel one-stage diffusion model designed to mitigate detail loss in fusion results caused by multi-step noise prediction. RED leverages supervision from the source images to enhance the stability of the diffusion process in image fusion tasks.
    % .(思想 为什么要reversible haochu 对于现有的方法的好处，具体的东西 ）
    % \item We introduce the reversible fusion strategy (RFS) to enable refined and efficient fusion, while retaining the ability to decompose fused outputs back into their source modalities, facilitating information retention.
    % \item We introduce the reversible fusion paradigm to enable the RED efficient training in custom-grad GPUs, while retaining the ability to decompose fused outputs back into their source modalities, facilitating information retention.
    % % (基于这个思想，我们提出了---instead 其他缺点）（解释为什么要怎么做）
    % \item We design and integrate a Prior-Guidance Module that incorporates prior information to refine and optimize the fused images. Extensive experiments validate the effectiveness and generalizability of RED across both visible-infrared and medical image fusion tasks.
    % （三个点的联系性）reversible 和diffusion的连接。）（为什么需要之间约束图像而不是噪声，）
    % 逆变换的可视化。体现可分解的能力。从底层机理上，为什么需要具有可以分解的能力，
    % 知识嵌入的好处，
    % 缺少一个独特的实验（类似于适应暗光修复）
    % 证明可分解的 解释性。
    % 补充材料。
    % 超参分析。
    % backbone
\end{itemize}

\section{Related Work}
\noindent\textbf{Image Fusion.}
End-to-end image fusion significantly enhances performance by learning nonlinear mapping relationships between source images. Early approaches primarily employed CNN-based models, which extract local features through hierarchical convolutional layers and integrate multi-source information by designing specific fusion strategies~\cite{ram2017deepfuse, li2018densefuse, li2021rfn, zhang2021sdnet, zhang2020ifcnn}. Transformer-based methods demonstrate superior performance due to the benefits of self-attention mechanisms, making them suitable for image fusion tasks~\cite{ma2022swinfusion, wang2022swinfuse}, leveraging transformer blocks to capture global dependencies. However, these models are often computationally intensive. To further improve both local and global feature extraction, hybrid CNN-Transformer architectures have been proposed, combining to capture low-level details and  high-level semantic~\cite{zhao2023cddfuse, yuan2022multimodal}. In parallel, state space models such as the Mamba ~\cite{gu2023mamba} offer promising solutions for modeling long sequences with linear complexity, and have been successfully applied to image fusion tasks, delivering high efficiency at low computational cost~\cite{xie2024fusionmamba, peng2024fusionmamba, li2024mambadfuse}. However, these models typically operate in a single training and inference stage, which may limit their ability to fully capture relevant information from source images. Our RED employs an iterative fusion strategy, progressively incorporating new details at each step to enable a more thorough and refined fusion process.

\noindent\textbf{Diffusion-based Image Fusion.}
Diffusion models have recently emerged as powerful generative frameworks. The DDPM~\cite{ho2020denoising} formulates a Markov chain that gradually perturbs data into noise and then reconstructs it via iterative sampling. However, accurate noise estimation typically requires a large number of sampling steps, making inference costly and potentially destabilizing consistency. To mitigate this, more efficient and consistency-preserving samplers such as DDIM~\cite{song2020denoising} and the Improved Diffusion framework~\cite{nichol2021improved} reduce the original thousands of steps to fewer than a hundred.
The Latent Diffusion Model (LDM)~\cite{rombach2022high} further cuts computation by performing diffusion in a compressed latent space. Nevertheless, the standard DMs training objective focuses on noise prediction for diversity image synthesis, rather than explicitly enforcing the structural fidelity and cross-modal consistency required by image fusion.
To address this gap, we propose a new diffusion-based paradigm, RED, tailored to fusion requirements, which optimizes the diffusion backbone end-to-end with explicit fusion constraints and produces fused images directly, aligning the generative objective with modality alignment and structure preservation while maintaining efficiency.

\noindent\textbf{Reversible Neural Networks.}
Reversible neural networks are a class of architectures capable of mapping observations directly to their underlying true states. NICE~\cite{dinh2014nice}, a generative model, learns a nonlinear deterministic transformation of the data, effectively mapping latent variables to a factorized distribution. RevNet~\cite{gomez2017reversible}, a variant of ResNet, achieves comparable accuracy while reducing memory usage by storing intermediate values only during backpropagation. Prior studies~\cite{genzel2022solving, ardizzone2018analyzing} have shown that reversible neural networks serve as powerful analytical tools for identifying multi-modal structures in parameter space, uncovering parameter correlations, and detecting unrecoverable parameters. This memory-efficient architecture makes reversible networks particularly well-suited to generative modeling, as normalizing flows enable exact, likelihood-based training.
\section{Preliminary}
% \subsection{Reversible Neural Network}
% \subsection{Reversible Neural Network}
\textbf{DDIM.} The denoising diffusion implicit model (DDIM)~\cite{song2020denoising} proposed a more efficient class of iterative sampling based on DDPM~\cite{ho2020denoising}:
 \begin{equation}
     x_{t-1}=\sqrt{\bar{\alpha}_{t-1}} x_0+\sqrt{1-\bar{\alpha}_{t-1}}\epsilon_t
 \end{equation}
% where $x_0=\frac{x_t-\sqrt{1-\bar{\alpha}_t}\epsilon_t}{\sqrt{\bar{\alpha}_t}}$. 
where $x_0=(x_t-\sqrt{1-\bar{\alpha}_t}\epsilon_t)/\sqrt{\bar{\alpha}_t}$ denotes the current denoised image. This diffusion-based framework unifies principles from hierarchical variational inference and score-based generative modeling, offering stable training and high-quality sample generation.

\noindent\textbf{Reversible Residual Block.} RevNets~\cite{gomez2017reversible} can be constructed using reversible residual blocks, which are designed to be mathematically invertible, thereby reducing memory consumption during training. Building on this concept, the Re$^2$TAL~\cite{zhao2023re2tal} further enhances reversibility and computational efficiency by rewiring connections, enabling the lightweight fine-tuning of consecutive residual modules within the same stage. These modules are rewired to ensure reversibility, with each residual connection skipping an additional block, denoted as $F$ and $G$. The following equation formally describes the rewiring process:
\begin{equation}
\left\{
\begin{array}{l}
y_1 = x_0 \\
x_1 = F(x_0) + y_0
\end{array}
\right.
\quad \Rightarrow \quad
\left\{
\begin{array}{l}
y_2 = x_1 \\
x_2 = G(x_1) + y_1
\end{array}
\right.
\label{eq:eq4}
\end{equation}
These equations are reversible, allowing recovery of inputs $x_1$, $y_1$ from $x_2$, $y_2$, and subsequently $x_0$, $y_0$ from $x_1$, $y_1$. The corresponding inverse computation is given by:
\begin{equation}
    \left\{
\begin{array}{l}
x_0 = y_1 \\
y_0 = x_1 - F(x_0)
\end{array}
\right.
\quad \Leftarrow \quad
\left\{
\begin{array}{l}
x_1 = y_2 \\
y_1 = x_2 - G(x_1)
\end{array}
\right.
\label{eq:eq5}
\end{equation}
This reversible structure allows efficient memory usage during training by eliminating the need to store intermediate activations, making it particularly suitable for deep and resource-constrained networks.

\section{Method}

\begin{figure*}
\centering
\includegraphics[width=1\linewidth]{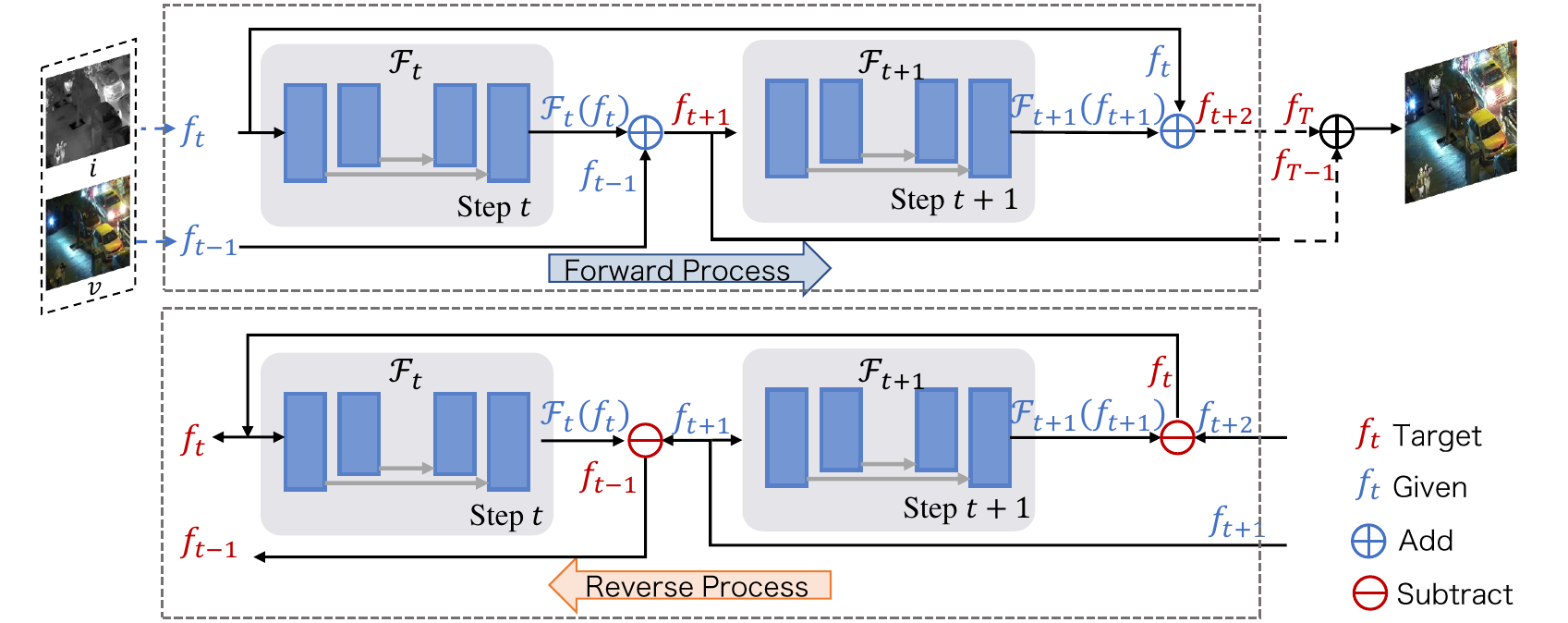}%%%%%%%%%%%%%%%%scale=缩小比例，或者用width=2in
\vspace{-5pt}
\caption{Workflow of the proposed RED model. The illustration depicts the reversible diffusion process and the knowledge-guidance module. The backbone adopts a U-Net architecture, where the standard residual blocks are replaced by reversible residual blocks.}
\vspace{-6pt}
\label{fig:frame}
\end{figure*}

The goal of image fusion is to generate a fused image $f$ from visible $v$ and infrared $i$ images, where $i,v,f\in \mathcal{R}^{C\times H\times W}$ $H$ and $W$ represent the height and width, and $C$ signifies the channels.

\subsection{End-to-end Diffusion Model for Image Fusion}
We propose an end-to-end training strategy to address the inherent ambiguities in measurements and the task gap present in the diffusion process. To this end, we introduce a novel diffusion-based image fusion framework that reformulates the DDIM~\cite{nichol2021improved} sampling procedure as a $T$-layer neural network, denoted as $\mathcal{F} = \mathcal{F}_1 \circ \mathcal{F}_2 \circ \cdots \circ \mathcal{F}_T$, where each layer $\mathcal{F}_t$ represents a sequential sampling step. Moreover, the coefficients ${\alpha}_{t=1}^T$ are jointly fine-tuned~\cite{chen2025invertible}. This layered structure enables progressive fusion, allowing the model to iteratively refine the fused image. To improve training efficiency, we adopt Latent Diffusion Models (LDMs) as the backbone and retain only the outermost two layers. The up-sampling and down-sampling operations, such as pixel-shuffle and unpixel-shuffle, are used to replace the role of the variational autoencoder (VAE). The framework is trained to minimize the discrepancy between the estimated output $f = \mathcal{F}(x, y)$ and the ground truth $(x, y)$ using standard loss functions commonly applied in image fusion tasks. This design enhances adaptability and overall performance, enabling the framework to generalize across diverse fusion scenarios.

\subsection{Reversible Image Fusion}
To implement end-to-end diffusion, we designed a reversible image fusion framework for efficient fusion. Inspired by~\cite{zhao2023re2tal}, we reformulate the rewiring process for image fusion, enabling progressive fusion. The coupling blocks consist of a diffused process with $f=\mathcal{F}(x,y)$, the inverse is the $[x,y]=\mathcal{F}^{-1}(f)$. In the following, we mathematically formulate the reversible modules. Firstly, we propose wiring to enable invertibility by introducing new connections into the sampling steps. Each input of the sampling step $\mathcal{F}_t$ is connected with the output of the $\mathcal{F}_{t-1}$. To represent the backpropagation process of the reverse process, it is beneficial to rephrase the forward (left) and reverse (right) computations in the following manner.  
% To better represent the backpropagation process of the reverse process, we introduce an intermediate variable called $z$,
\begin{equation}
% \text{Forward}
\left\{
\begin{array}{l}
f_{t+1} = f_{t-1}+\mathcal{F}_{t}(f_t)   \\
f_{t-1} = f_{t} \\
f_{t} = f_{t+1} \\
\end{array}
\right.
% \quad \Leftrightarrow \quad
\Leftrightarrow
% \text{Reverse}
\left\{
\begin{array}{l}
f_{t} = f_{t-1} \\
f_{t+1} = f_{t}\\ 
f_{t-1} = f_{t+1} -\mathcal{F}_{t}(f_t)\\
\end{array}
\right.
\label{eq:eq6}
\end{equation}

% \begin{equation}
% \text{Forward}
% \left\{
% \begin{array}{l}
% f_{t+1} = f_{t-1}+\mathcal{F}_{t}(f_t)   \\
% f_{t-1} = f_{t} \\
% f_{t} = f_{t+1} \\
% \end{array}
% \right.
% \quad \Leftrightarrow \quad 
% \text{Reverse}
% \left\{
% \begin{array}{l}
% f_{t} = f_{t-1} \\
% f_{t+1} = f_{t}\\ 
% f_{t-1} = f_{t+1} -\mathcal{F}_{t}(f_t)\\
% \end{array}
% \right.
% \label{eq:eq6}
% \end{equation}

% \begin{equation}
% \begin{gathered}
% \text{Forward}\ \left\{
% \begin{array}{l}
% f_{t+1} = f_{t-1}+\mathcal{F}_{t}(f_t) \\
% f_{t-1} = f_{t} \\
% f_{t}   = f_{t+1}
% \end{array}\right.
% \\[6pt]
% \Updownarrow
% \\[6pt]
% \text{Reverse}\ \left\{
% \begin{array}{l}
% f_{t}   = f_{t-1} \\
% f_{t+1} = f_{t} \\
% f_{t-1} = f_{t+1} - \mathcal{F}_{t}(f_t)
% \end{array}\right.
% \end{gathered}
% \label{eq:eq6}
% \end{equation}
Note that, even though the $f_{t-1}=f_t$ in Eq.~\ref{eq:eq6}, the two variables represent distinct nodes of the computation graph, so the total derivatives $f_{t-1}$ and $f_t$ are different.

In the forward process, we cache the final output of wired layers and the intermediate features are freed up. By iteratively applying Algorithm~\ref{alg:alg_red}, backpropagation can be performed across a sequence of reversible blocks using only the activations and their derivatives from the topmost layer. The complete diffusion process becomes reversible when multiple reversible modules are stacked consecutively in a network (Fig.~\ref{fig:frame}). In this configuration, we execute the backpropagation from $(f_0, f_1)$ back to $(i, v)$, reconstruction begins at the final module, with each module's input sequentially recovered using Eq.~\ref{eq:eq4}. After computing the loss function $\mathcal{L}$. The final fusion output $f = wf_T+(1-w)f_{T-1}$, here the $w$ is a learnable parameter.
Note that this approach can be applied to arbitrary diffusion models without the need to design new neural networks. It significantly saves memory, time, and computation, improving reconstruction performance.

The U-Net architecture~\cite{ronneberger2015u} is commonly used as a noise estimator in diffusion models. It includes a series of down-sampling and mirrored up-sampling operations with skip connections. These blocks consist of residual and attention blocks, which are memory-intensive. To further improve memory efficiency, we extend the concatenation of the reversible residual block to the U-Net blocks at each step, as in RevNet~\cite{gomez2017reversible}, as shown in Eq.~\ref{eq:eq4} and Eq.~\ref{eq:eq5}. We group and concatenate consecutive transformation blocks within each block, making them reversible. During training, we clear all inputs and intermediate activations of these grouped blocks from memory, retaining only the features of the first and last convolutions and skip branches for backpropagation. This reversible design reduces memory usage and enables our method to train on consumer GPUs~\cite{chen2025invertible}.

\subsection{Loss Function}
As an end-to-end (E2E) framework, the loss functions used to train RED include SSIM loss, MAE loss, and maximum gradient loss. Detailed, the SSIM loss encourages the model to preserve important structural features, leading to more perceptually realistic results. It can be formulated as:$\mathcal{L}_{\text{SSIM}} = 2 - (\text{SSIM}(i,f)+\text{SSIM}(v,f)).$
The MAE loss aims to minimize the pixel-level differences and improve the visual similarity between the source images and the fused image denoted as:$\mathcal{L}_1 = ||i-f||_1 + ||v-f||_1$.
% \begin{equation}
%     \mathcal{L}_1 = ||i-f||_1 + ||v-f||_1.
% \end{equation}
The maximum gradient loss helps improve the perceptual quality of images by focusing on maintaining edge sharpness. It can be defined as:$\mathcal{L}_{grad} = \frac{1}{HW}||\nabla f -max(\nabla v, \nabla i)||_1$, here $\nabla$ represents the Sobel operator, which extracts edge information from the image.
% \begin{equation}
%     \mathcal{L}_{grad} = \frac{1}{HW}||\nabla f -max(\nabla v, \nabla i)||_1,
% \end{equation}
The total loss function is defined as $\mathcal{L} = \mathcal{L}_{\text{SSIM}} + \mathcal{L}_{1} + \mathcal{L}_{\text{grad}}$. This formulation provides explicit supervision from the source images and the fused image, thereby enhancing the fidelity of the fusion results.
\begin{table*}
\centering
\footnotesize
\setlength{\tabcolsep}{3pt}
\renewcommand\arraystretch{1.2}
\caption{Quantitative comparison with SOTAs. The \textbf{bold}/\underline{underline} indicates the best and runner-up.}
\fontsize{10}{10}\selectfont\setlength{\tabcolsep}{1.2mm}
\scalebox{0.9}{
\begin{tabular}{lccccc|ccccc}
\toprule
\multicolumn{6}{c|}{Visible-Infrared Image Fusion on LLVIP Dataset} &  \multicolumn{5}{c}{Visible-Infrared Image Fusion on MSRS Dataset} \\ 
 & EI & AG & SF & $Q^{AB/F}$  & VIFF  &  EI   & AG & SF &  $Q^{AB/F}$  & VIFF  \\
\midrule 

UMF-CMGR~\cite{wang2022unsupervised} & 8.43 & 3.40 & 11.89 & 0.31 & 0.39
% & UMF-CMGR~\cite{wang2022unsupervised} 
& 6.32 & 2.41 & 7.47 & 0.26 & 0.41
\\
YDTR~\cite{tang2022ydtr} 
& 8.91 & 3.72 & 13.82 & 0.35 & 0.47 
% & YDTR~\cite{tang2022ydtr} 
& 6.55 & 2.48 & 7.72 & 0.35 & 0.57
\\
DeFusion~\cite{liang2022fusion}
& 9.94 & 3.87 & 12.02 & 0.42 & 0.58
% & DeFusion~\cite{liang2022fusion}
& 7.13 & 2.66 & 8.05 & 0.45 & 0.58
\\
U2Fusion~\cite{xu2020u2fusion} 
& 11.48 & 4.53  & 14.25 & 0.43  & 0.48
% & U2Fusion~\cite{xu2020u2fusion} 
& 7.78 & 2.96 & 8.27 & 0.15 & 0.17
\\
CDDFuse~\cite{zhao2023cddfuse}
& 13.31 & 5.40 & 18.50 & 0.58 & 0.66
% & CDDFuse~\cite{zhao2023cddfuse}
& 10.07 & 3.78 & 11.57 & 0.58 & 0.72
\\
DDFM~\cite{zhao2023ddfm}
& 8.10 & 3.32 & 11.89 & 0.35 & 0.59
% & DDFM~\cite{zhao2023ddfm}
& 5.37 & 2.04 & 6.44 & 0.29 & 0.60
\\ 
TC-MoA~\cite{zhu2024task}
&\underline{14.18} & \underline{5.69} & 18.79 & 0.60 & \underline{0.71}
% & TC-MoA~\cite{zhu2024task}
& 8.55 & 3.15 & 9.19 & \underline{0.61} & \underline{0.83}
\\
TTD~\cite{cao2024test}
& 13.83 & 5.44 & \underline{19.18} & \underline{0.65} & 0.69
% & TTD~\cite{cao2024test}
& 9.71 & 3.71 & 11.53 & 0.55 & 0.66
\\
Text-DiFuse~\cite{zhang2024text}
& 12.56 & 4.85 & 15.53 & 0.40 & 0.52
% & Text-DiFuse~\cite{zhang2024text}
& \underline{10.36} & \underline{3.84} & 11.51 & 0.44 & 0.70
\\
% RED (Ours)
% & \textbf{14.74}\textcolor{Red}{\scriptsize(+0.56)} & \textbf{5.90} & \textbf{19.34} & \textbf{0.75} & \textbf{0.92}
% & \textbf{10.39} & \textbf{3.90} & \underline{11.57} & \textbf{0.70} & \textbf{0.98}
% \\

RED (Ours)
& \textbf{14.74}\textcolor{Red}{\scriptsize(+0.56)} & \textbf{5.91}\textcolor{Red}{\scriptsize(+0.22)} & \textbf{19.29}\textcolor{Red}{\scriptsize(+0.11)} & \textbf{0.74}\textcolor{Red}{\scriptsize(+0.09)} & \textbf{0.93}\textcolor{Red}{\scriptsize(+0.22)}
& \textbf{10.39}\textcolor{Red}{\scriptsize(+0.03)} & \textbf{3.90}\textcolor{Red}{\scriptsize(+0.06)} & \textbf{11.57}\textcolor{Red}{\scriptsize(-0.00)} & \textbf{0.70}\textcolor{Red}{\scriptsize(+0.09)} & \textbf{0.98}\textcolor{Red}{\scriptsize(+0.15)}
\\

\bottomrule

\multicolumn{6}{c|}{Visible-Infrared Image Fusion on M3FD Dataset} & \multicolumn{5}{c}{Medical Image Fusion on Harvard Dataset} \\ 
 & EI & AG & SF & $Q^{AB/F}$ & VIFF  &  EI   & AG  & SF  &  $Q^{AB/F}$  & VIFF   \\
\midrule 

UMF-CMGR~\cite{wang2022unsupervised}  
& 7.75 & 2.98 & 8.90 & 0.40 & 0.60 
% & UMF-CMGR~\cite{wang2022unsupervised} 
& 16.22 & 6.32 & 22.66 & 0.41 & 0.37 
\\
YDTR~\cite{tang2022ydtr} 
& 8.68 & 3.36 & 10.22 & 0.48 & 0.62
% & YDTR~\cite{tang2022ydtr} 
& 15.25 & 5.90 & 21.64 & 0.41 & 0.52
\\
DeFusion~\cite{liang2022fusion}
& 6.89 & 2.62 & 7.48 & 0.34 & 0.53 
% & DeFusion~\cite{liang2022fusion}
& 16.24 & 6.14 & 21.79 & 0.51 & 0.52
\\
U2Fusion~\cite{xu2020u2fusion}
& 8.17 & 3.11 & 9.12 & 0.38 & 0.49 
% & U2Fusion~\cite{xu2020u2fusion}
& 16.23 & 6.12 & 21.18 & 0.46 & 0.46
\\
CDDFuse~\cite{zhao2023cddfuse}
& 12.45 & 4.80 & 14.71 & 0.52 & 0.58
% & CDDFuse~\cite{zhao2023cddfuse}
& 19.53 & 7.44 & \underline{26.24} & 0.62 & \underline{0.67}
\\
DDFM~\cite{zhao2023ddfm}
& 8.17 & 3.11 & 9.12 & 0.38 & 0.49
% & DDFM~\cite{zhao2023ddfm}
& 13.96 & 5.40 & 20.76 & 0.35 & 0.43 
\\ 
TC-MoA~\cite{zhu2024task}
& 12.03 & 4.61 & 13.81 & 0.63 & 0.78
% & TC-MoA~\cite{zhu2024task}
& 16.52 & 6.23 & 21.21 & \underline{0.64} & \underline{0.53}
\\
TTD~\cite{cao2024test}
& \underline{12.88} & \underline{4.98} & \textbf{14.84} & 0.60 & 0.64
% & TTD~\cite{cao2024test}
& \underline{19.77} & \textbf{7.80} & \textbf{28.97} & 0.63 & 0.49
\\
Text-DiFuse~\cite{zhang2024text}
& 7.51 & 2.81 & 8.47 & 0.15 & 0.16 
% & Text-DiFuse~\cite{zhang2024text}
& 17.99 & 6.76 & 24.27 & 0.45 & 0.48
\\
% RED (Ours)
% & \textbf{13.04} & \textbf{5.00} & \undeline{14.81} & \textbf{0.70} & \textbf{0.93}
% % & Ours
% & \textbf{20.24} & \underline{7.67} & 26.14 & \textbf{0.71} & \textbf{0.69}
% \\
RED (Ours)
& \textbf{13.04}\textcolor{Red}{\scriptsize(+0.16)} & \textbf{5.00}\textcolor{Red}{\scriptsize(+0.02)} & \underline{14.81}\textcolor{Blue}{\scriptsize(-0.03)} & \textbf{0.70}\textcolor{Red}{\scriptsize(+0.07)} & \textbf{0.93}\textcolor{Red}{\scriptsize(+0.15)}
& \textbf{20.24}\textcolor{Red}{\scriptsize(+0.47)} & \underline{7.67}\textcolor{Blue}{\scriptsize(-0.13)} & 26.14\textcolor{Blue}{\scriptsize(-1.83)} & \textbf{0.71}\textcolor{Red}{\scriptsize(+0.07)} & \textbf{0.69}\textcolor{Red}{\scriptsize(+0.02)}
\\
\bottomrule

\end{tabular}
}
\vspace{-8pt}
\label{tab:comparison}
\end{table*}
\section{Experiments}
\subsection{Setup}
\label{setting}
\noindent\textbf{Implementation Details.}
We conduct experiments on three image fusion datasets: LLVIP~\cite{jia2021llvip}, MSRS~\cite{tang2022piafusion}, and M$^3$FD~\cite{liu2022target}, using PyTorch on a machine equipped with two NVIDIA GeForce RTX A6000 GPUs. Additionally, we perform medical image fusion experiments on the Whole Brain Atlas dataset from Harvard Medical School~\cite{HarvardMedical} as a complementary evaluation.
The model is trained on the LLVIP training set and evaluated following the TC-MoA~\cite{zhu2024task}. To assess the generalization ability, the model trained on LLVIP is directly evaluated on the MSRS, M$^3$FD, and Harvard Medical datasets without fine-tuning, using the same evaluation metrics.
During training, we set $T=2$, image pairs are randomly cropped into $128 \times 128$ patches, with a batch size of 4. The model is optimized using the Adam optimizer with parameters $\beta_1 = 0.9$ and $\beta_2 = 0.999$.

% \noindent\textbf{Datasets.} 
% For multi-modal image fusion task, we conducted experiments on the LLVIP~\cite{jia2021llvip} dataset and referred to the test set outlined in the work by Zhu et al.~\cite{zhu2024task}.

\noindent\textbf{Evaluation Metrics.} 
We evaluated the fusion results in both quantitative and qualitative. Qualitative evaluation is primarily based on subjective visual assessments conducted by individuals. We expect the fused image to exhibit rich texture details and abundant color saturation.
Objective evaluation focuses primarily on measuring the quality assessments of individual fused images and their deviations from the source images. For different task scenarios, the same evaluation metrics are used, specifically, we employ five metrics including edge intensity (EI), average gradient (AG), spatial frequency (SF), gradient-based similarity measurement ($Q^{AB/F}$) and visual information fidelity for fusion (VIFF) for evaluation metrics.

\begin{figure*}
\centering
\includegraphics[width=1\linewidth]{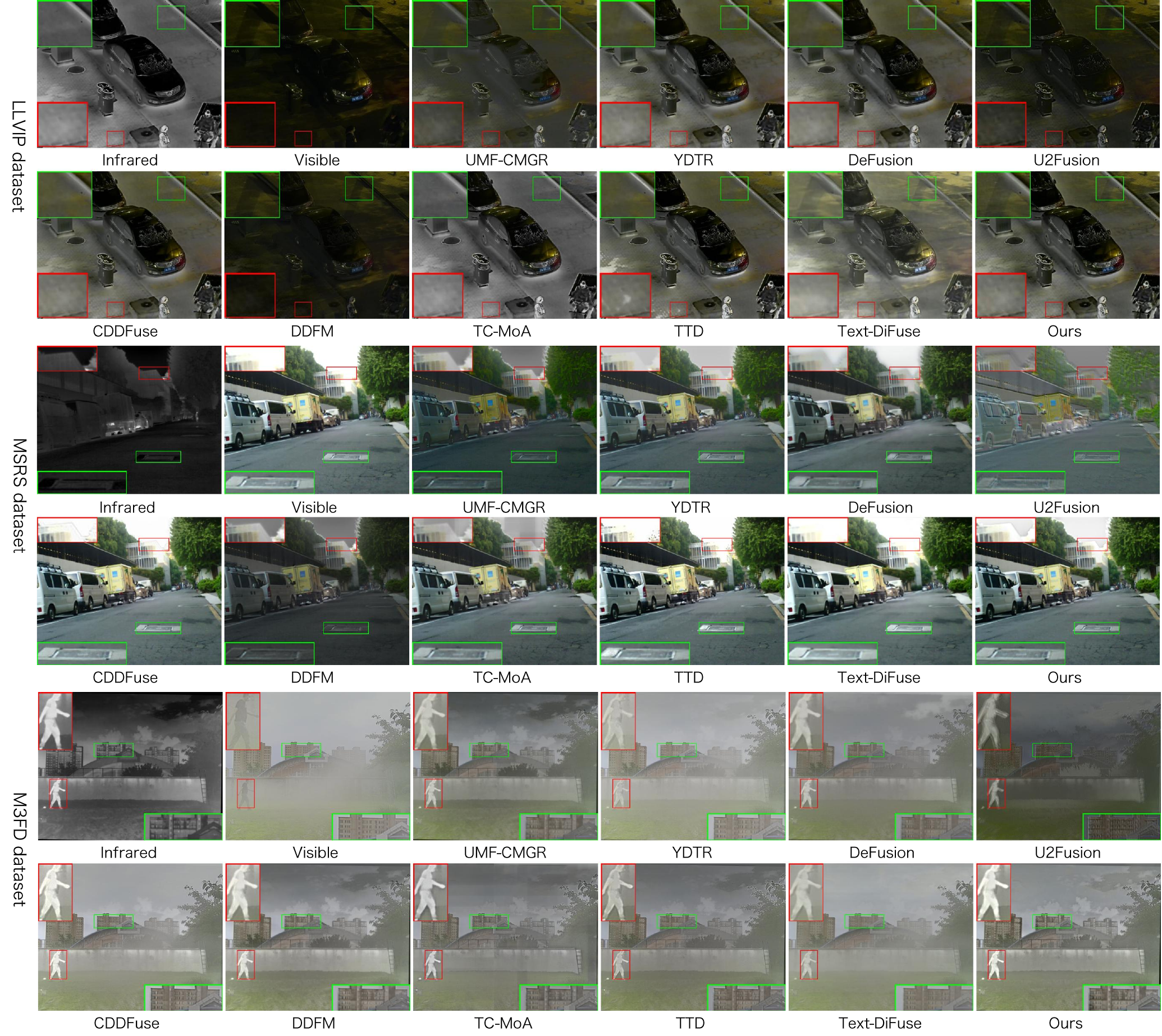}%%%%%%%%%%%%%%%%scale=缩小比例，或者用width=2in
\vspace{-5pt}
\caption{Qualitative comparisons of SOTA methods in the LLVIP, MSRS, M$^3$FD  datasets.}
\vspace{-10pt}
\label{fig:comparison123}
\end{figure*}
\subsection{Comparison with SOTA methods}

\noindent\textbf{Competing Methods.} 
We compared our method with seven recent competing methods, including U2Fusion~\cite{xu2020u2fusion}, UMF-CMGR~\cite{wang2022unsupervised}, YDTR~\cite{tang2022ydtr}, DeFusion~\cite{liang2022fusion}, CDDFuse~\cite{zhao2023cddfuse}, DDFM~\cite{zhao2023ddfm},  TC-MoA~\cite{zhu2024task}, TTD~\cite{cao2024test}, and Text-DiFuse~\cite{zhang2024text}.

\noindent\textbf{Quantitative Comparisons.} As shown in Table~\ref{tab:comparison}, the evaluation metrics demonstrate that our fusion results consistently outperform existing methods. Our approach excels by ranking first or second in all evaluated metrics, showcasing remarkable performance across various VIF datasets. Notably, the competitive values in the EI, AG and SF metrics reflect the superior detail preservation and enhanced texture quality achieved by RED. The highest values of $Q^{AB/F}$ and VIFF also highlight that our fusion outputs exhibit the highest correlation with human visual perception, underscoring their closer alignment with human perception. This ensures that the generated images are not only perceptually accurate but also retain the integrity of the original source information. The outperformance across diverse datasets reinforces the effectiveness and robustness of RED in complex application scenarios, indicating that comprehensive information has been preserved. Meanwhile, MIF achieves the best performance in EI, $Q^{AB/F}$, and VIFF, as well as the second-best result in AG, demonstrating RED’s adaptability across different scenarios and tasks to produce high-quality results.

\begin{figure*}
\centering
\includegraphics[width=1\linewidth]{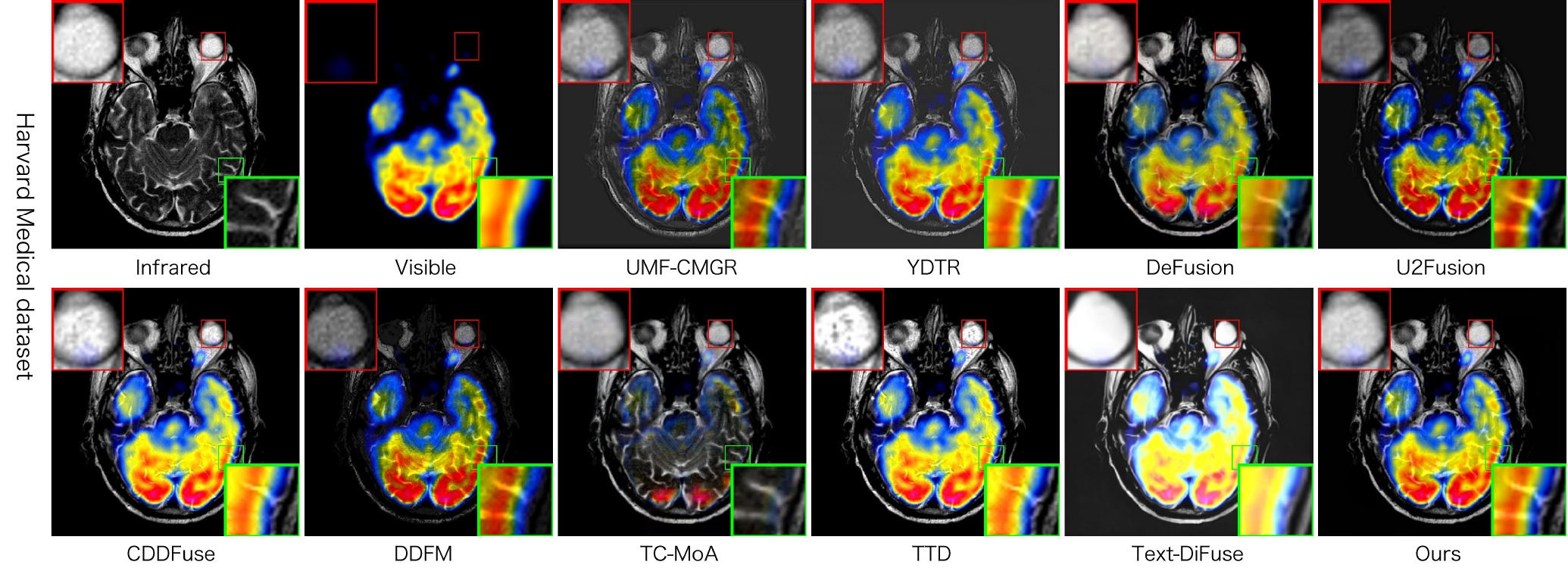}%%%%%%%%%%%%%%%%scale=缩小比例，或者用width=2in
% \vspace{-15pt}
\caption{Qualitative comparisons of SOTA methods in Harvard Medical dataset.}
\vspace{-6pt}
\label{fig:comparison_4}
\end{figure*}

\noindent\textbf{Qualitative Comparisons.} Qualitative comparison results with state-of-the-art (SOTA) image fusion methods are shown in Fig.\ref{fig:comparison123}. In rows 1–2 of Fig.\ref{fig:comparison123}, our method effectively preserves shadow details of the tree in the dark visible-light regions while simultaneously retaining the texture of the floor tiles in the infrared image. This is particularly evident when compared to Text-DiFuse, demonstrating the effectiveness of our image-level constraints.  In the red box of In rows 3–4 of Fig.\ref{fig:comparison123}, highlighted in the red box, our method excels in retaining more infrared information in overexposed sky regions, where other methods typically lose detail. This dual preservation of information is critical for maintaining scene context in both modalities benefited from the iterative fusion process, which enables RED to adapt effectively to complex scenarios. In Fig.~\ref{fig:comparison123} row 5-6, our approach demonstrates robustness in the presence of smoke, resulting in fused images with superior clarity and contrast compared to other methods. Moreover, RED achieves the best results in the MIF scenario, showing the highest contrast and most distinct structural features shown in Fig.~\ref{fig:comparison_4}, which highlights its strong task adaptability. 
% Overall, our method significantly outperforms other fusion techniques, achieving a more effective integration of infrared and visible-light information. This performance underscores the strength of RED in preserving semantic details, maintaining fidelity, and aligning well with human visual perception.

% Furthermore, the incorporation of the basic condition and enhanced conditions enables effective preservation of the background and texture. This comparison underscores our model's efficacy in image fusion, resulting in outstanding visual outcomes. As shown in Fig.~\ref{fig:comparison_1}, our method showcases superior visual quality compared to other approaches. Specifically, our method excels in preserving intricate texture details well lid in low light (Fig.~\ref{fig:comparison_1} red box). Although TC-MoA and MUFusion approach our method in retaining details, they exhibit visible artifacts, blur, and low contrast—characteristics absent in CCF (Fig.~\ref{fig:comparison_1}, green box). CCF exhibits the highest contrast, the clearest details, and the most information content, further highlighting its superiority in preserving texture details. Its excellent detail retention and clear background generation further demonstrate the effectiveness of our proposed method. 

\subsection{Ablation Study}
\textbf{Qualitative and Quantitative Analysis.}We conduct a series of ablation studies on the LLVIP dataset to evaluate the effectiveness of the proposed modules in RED. Table~\ref{tab:abla} presents several variants of RED, each with a specific module ablated. Specifically, Reverse \uppercase\expandafter{\romannumeral1} denotes the reversible fusion strategy, Reverse \uppercase\expandafter{\romannumeral2} refers to the reversible residual block, and without DDPM indicates the exclusion of the denoising process.
Without the reversible fusion strategy, the model runs with huge memory during training, indicating that this component is essential for enabling end-to-end training under limited memory budgets. This confirms its critical role in reducing memory consumption via reversible computation. When Reverse \uppercase\expandafter{\romannumeral1} is included but Reverse \uppercase\expandafter{\romannumeral2} is removed, the model achieves a competitive PSNR of 19.35 and EI of 14.71, highlighting that reversible residual blocks are essential for structural fidelity and perceptual quality. Adding Reverse \uppercase\expandafter{\romannumeral2} results in minimal impact on memory usage, demonstrating the strong contribution to visual consistency. However, the most significant improvement is achieved when all components, including the denoising diffusion process, are activated. In this configuration, RED achieves the best overall performance with a PSNR of 19.29, AG of 5.91, EI of 14.74 and the highest VIFF of 0.93, indicating enhanced information preservation. Furthermore, we provide more ablation analysis in the Appendix. These all validate the effectiveness of each proposed component. Overall, RED achieves superior performance with substantial memory efficiency, underscoring the efficacy of our architectural design.

% \begin{wraptable}{r}{0.6\textwidth}
\begin{table}
\centering
% \vspace{-20pt}
\tabcolsep=2pt
\footnotesize
\caption{Ablation studies on the LLVIP dataset. \textbf{Bold} indicates the best performance. "Mem." denotes memory usage, and "Inf." refers to inference time. OOM indicates out-of-memory.}
\fontsize{9}{10}\selectfont\setlength{\tabcolsep}{0.6mm}
\vspace{-5pt}
\scalebox{0.75}{
\begin{tabular}{ccc|ccccccc}
\toprule
Reverse \uppercase\expandafter{\romannumeral1} & Reverse \uppercase\expandafter{\romannumeral2} & DDIM & EI   & AG  & PSNR &  $Q^{AB/F}$  & VIFF & Mem. (GB)& Inf. (s)\\
\midrule  
 -& $\checkmark$  &$\checkmark$ & 14.44 & 5.79 & 19.01 & 0.73 & 0.91 & 20.1 & 1.09 \\
 
 % $\checkmark$ &   - &$\checkmark$ & $14.29$&$5.63$&$17.86$&$0.72$&$0.82$ & 7.10& 0.95 \\
$\checkmark$ & -  &$\checkmark$ & $14.71$ & $5.90$ & $\mathbf{19.35}$ & $0.74$ & $0.92
$ & $9.44$ & $1.09$\\
$\checkmark$ & $\checkmark$  &-& $14.47$ & $5.80$ & $19.07$ & $0.73$ & $0.92$ & $9.44$ & $1.10$\\
$\checkmark$ & $\checkmark$ &$\checkmark$ & $\mathbf{14.74}$ & $\mathbf{5.91}$ & $19.29$ & $\mathbf{0.74}$ & $\mathbf{0.93}$  & $7.14$ & $1.30$
\\			
\bottomrule
\end{tabular}
}
\vspace{-3pt}
\label{tab:abla}
% \end{wraptable}
\end{table}

\begin{figure}
\centering
\includegraphics[width=1\linewidth]{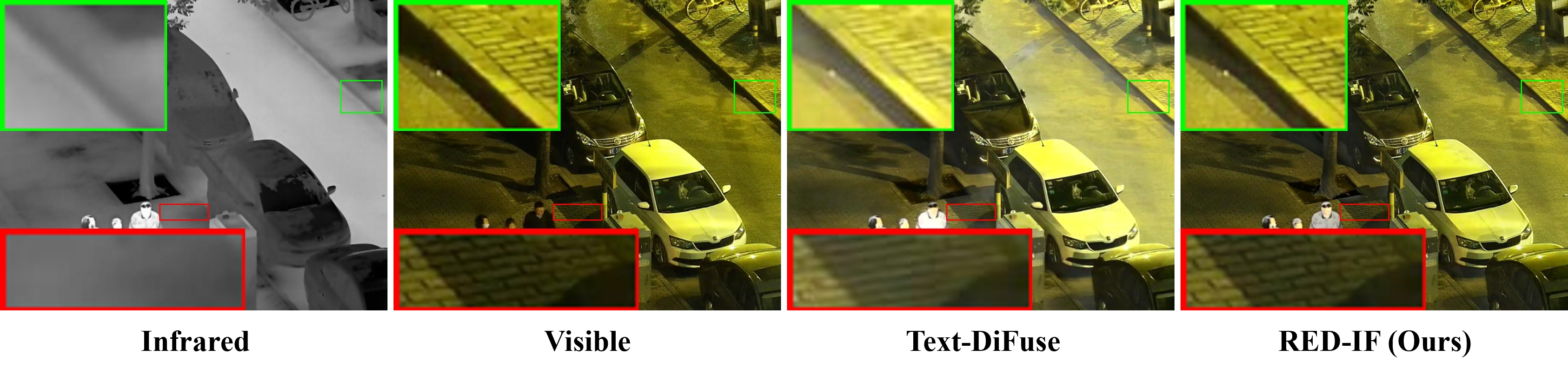}%%%%%%%%%%%%%%%%scale=缩小比例，或者用width=2in
\vspace{-10pt}
\caption{Visualization of representative details lost exhibited by Text-DiFuse.}
\vspace{-10pt}
\label{fig:dis_end}
\end{figure}
The RED model applies direct supervision to both the source images and the fused output, resulting in impressive performance. But why does image-level supervision lead to such encouraging fusion results? The key lies in the explicit alignment between input modalities and the target image, which helps preserve fine details and semantic consistency.
In contrast, diffusion models trained with weak or implicit supervision tend to interpolate smoothly between data modes seen during training~\cite{mei2025improving}, which can lead to over-smoothed outputs and a loss of fine structures. Moreover, recent studies have shown that minor inconsistencies in conditions can be progressively amplified during iterative sampling~\cite{aithal2024understanding}, leading to the accumulation of artifacts and deviations from real data distribution. These behaviors are particularly problematic in Markovian diffusion processes, where each step depends solely on the previous state, exacerbating error accumulation.
Such issues are clearly observed in Text-IF, as illustrated in Fig.~\ref{fig:dis_end}, which demonstrates both detail degradation (green box) and structural distortions (red box). By contrast, RED benefits from strong image-level supervision, which anchors the fusion process to ground-truth semantics and structures, thereby mitigating error propagation and yielding more refined and faithful fusion results.
%% why end-to-end superise 
% \begin{wrapfigure}{r}{0.5\textwidth}  % 右侧插入图片，占用一半的宽度
%     \centering
%     \includegraphics[width=0.48\textwidth]{images/fig_discuss_1.png}  % 调整图片大小
%     \caption{discuss} 
% \label{fig:dis_end}
% \end{wrapfigure}

\begin{figure*}
\centering
\includegraphics[width=1\linewidth]{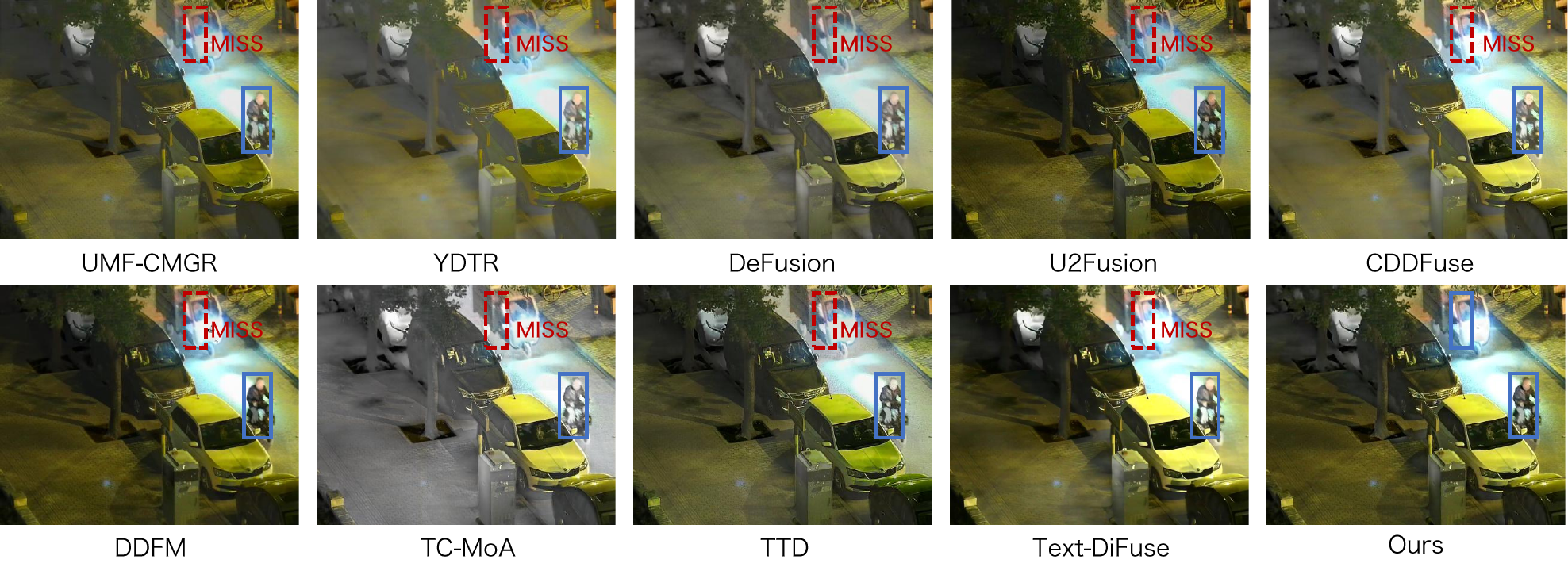}%%%%%%%%%%%%%%%%scale=缩小比例，或者用width=2in
\vspace{-10pt}
\caption{Visualization comparing SOTA methods on the object detection task.}
\vspace{-8pt}
\label{fig:fig_det}
\end{figure*}

\subsection{Downstream VIF Application}

\begin{table}
% \vspace{-20pt}
% \tabcolsep=4.2pt
\footnotesize
\centering
\caption{Comparison of object detection on LLVIP dataset. \textbf{Blod} indicates the best.}
\vspace{-5pt}
\fontsize{9}{10}\selectfont\setlength{\tabcolsep}{1.2mm}
\scalebox{1}{
\begin{tabular}{l|cccc}
\toprule
Method & Precision & Recall & mAP@.5 & mAP@.5:.95 \\
\midrule 
U2Fusion~\cite{xu2020u2fusion}  & $0.938$ & $0.898$ & $0.954$ & $0.631$ \\

UMF-CMGR~\cite{wang2022unsupervised}  & $0.946$ & $0.886$ & $0.950$ & $0.637$ \\

YDTR~\cite{tang2022ydtr}  & $0.946$ & $0.905$ & $0.958$ & $0.631$ \\

DeFusion~\cite{liang2022fusion}  & $0.945$ & $0.905$ & $0.959$ & $0.629$ \\

CDDFuse~\cite{zhao2023cddfuse} & $0.948$ & $0.899$ & $0.955$ & $0.631$ \\

DDFM~\cite{zhao2023ddfm} & $0.952$ & $0.904$ & $0.960$ & $0.630$ \\

TC-MoA~\cite{zhu2024task} & $0.939$ & $0.914$ & $0.961$ & $0.637$ \\

TTD~\cite{cao2024test}& $0.944$ & $0.894$ & $0.952$ & $0.630$ \\

Text-DiFuse~\cite{zhang2024text} & $0.937$ & $0.874$ & $0.939$ & $0.596$ \\

Dream-IF (Ours) & $\mathbf{0.949}$ & $\mathbf{0.906}$ & $\mathbf{0.962}$ & $\mathbf{0.639}$ \\
\bottomrule
\end{tabular}
}
\vspace{-10pt}
\label{tab:det}
\end{table}

In this section, we provide an evaluation of the downstream performance on the LLVIP dataset, taking object detection as an example, utilizing the YoloV11~\cite{khanam2024yolov11} detection backbone. The model is trained for a total of 200 epochs using the SGD optimizer, with an initial learning rate set to 0.01. For each comparison method, we retrain the model by incorporating the fusion results obtained from their respective methods, ensuring a fair comparison across all approaches. The evaluation is conducted using a comprehensive set of metrics, including precision, recall, mean average precision at IoU 0.5 (mAP@0.5), and mean average precision at IoU 0.5:0.95 (mAP@0.5:95). The results of these evaluations are systematically presented in Table~\ref{tab:det}, indicating that RED consistently outperforms all other comparison methods across all evaluation metrics, particularly mAP@0.5:95 increased by 0.019 than the suboptimal. Fig.~\ref{fig:fig_det} presents a visualization in which RED successfully detects a person under strong illumination—an instance where other methods fail. Additional visualizations are provided in the appendix. This superior performance highlights RED’s effectiveness in enhancing detection tasks and demonstrates its robust capability to adapt to downstream applications.
\begin{figure}
    \centering
    \begin{minipage}{0.49\textwidth}
        \centering
        \includegraphics[width=\linewidth]{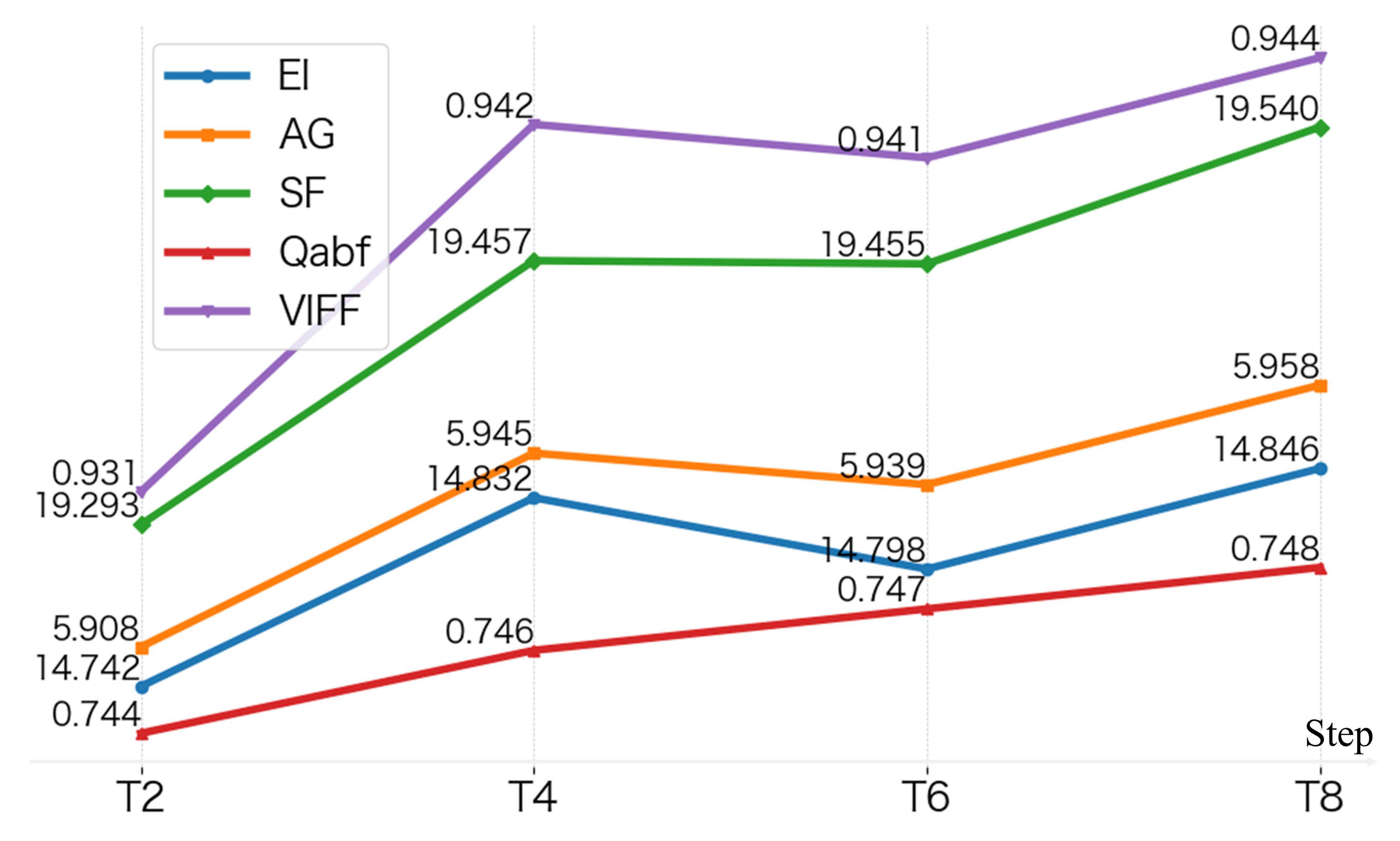}
                \vspace{-20pt}
        \caption{Quantitative results at different diffusion steps. Values are scaled for visualization.}
        \vspace{5pt}
        \label{fig:dis_step}
    \end{minipage}
    \hfill
    \begin{minipage}{0.49\textwidth}
        \centering
        \includegraphics[width=\linewidth]{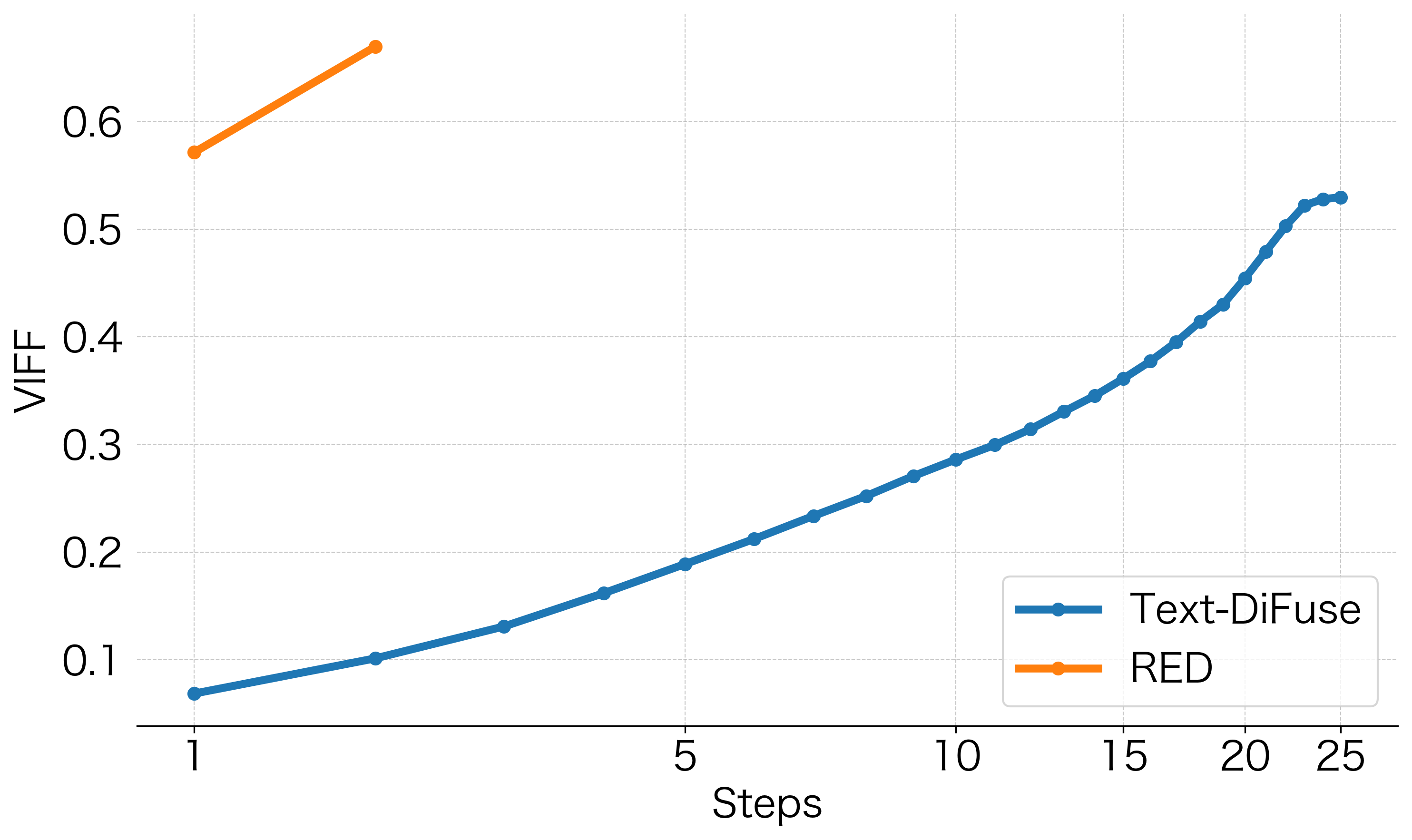}
            \vspace{-20pt}
        \caption{Comparison VIFF at different diffusion steps with Text-DiFuse.}
        \vspace{-15pt}
        \label{fig:fig_dis1}
    \end{minipage}
\end{figure}

\section{Discussion}
\textbf{How Much Iterative Training?} We further investigate the impact of the number of iterative steps on model performance, as shown in Fig.~\ref{fig:dis_step} and visualize the intermediate output in the Appendix. The results indicate that increasing the number of iterations generally enhances performance. However, the marginal gain diminishes with additional steps, suggesting a saturation effect. Moreover, computational resource consumption increases approximately linearly with the number of iterations, revealing a trade-off between performance and efficiency.
To better understand the advantages of end-to-end supervision in RED, we compare its VIFF scores with those of the diffusion-based image fusion method Text-DiFuse~\cite{zhang2024text} with randomly selecting a sample, as shown in Fig.~\ref{fig:fig_dis1}. RED achieves high fidelity even at early stages of the diffusion process, with performance improving steadily over subsequent iterations. In contrast, Text-DiFuse exhibits a slower and less efficient increase in fidelity, reflecting a more delayed fusion process.

By discarding the traditional Markov-based framework, RED introduces direct supervision between the source images and the fused output. This design enables the model to achieve high-quality results with fewer diffusion steps, making it more suitable for image fusion tasks that demand both accuracy and computational efficiency.

% how much iterative training 
% 中间结果的可视化 
% 为什么需要更少的步骤就可以实现 （中间可视化展示）
% sd单模态重建。我发现通过这样子一个步骤不需要叠加很多步数。在图像融合任务中，不需要很多步骤，就可以实现这样子的融合，如ddfm 100，text-difuse 25， ccf 100，(表格对比） 需要很多步长，但是我们只需要。 ccf或者ddfm的生成结果指标。
% 

\section{Conclusion}
In this paper, we propose an end-to-end diffusion model for image fusion that leverages strong supervision to mitigate the inconsistency generation in diffusion models. We employ a reversible architecture to enable end-to-end training while significantly reducing memory consumption. Furthermore, our exploration of iterative steps demonstrates the model’s capabilities and inspires a novel paradigm for image fusion. Empirical results show that RED outperforms competing methods, achieving superior performance across various datasets. In future work, we aim to develop automatic step selection strategies to balance computational cost and performance metrics, thereby enabling more practical and effective image fusion.
% CVPR 2026 Paper Template; see https://github.com/cvpr-org/author-kit

\clearpage
\setcounter{page}{1}
\maketitlesupplementary

\section{Technical Appendices and Supplementary Material}

\subsection{More Details of Reversible Fusion}
% 往前放，详细介绍。
% The complete RED algorithm is provided in Algorithm~\ref{alg:alg_red}, which outlines both the forward and backward processes in the reversible fusion module. This includes how intermediate states are recomputed during backpropagation to achieve memory efficiency.

Our RED is modality-agnostic. In this work, we instantiate it with two inputs by initializing $f_0 = v$ and $f_1 = i$, where $v$ and $i$ denote the visible and infrared images, respectively. The fused image is produced from the final states $f_T$ and $f_{T-1}$.
The reverse starts from $f_T$ and $f_{T-1}$ obtained in the forward trajectory. As detailed in Algorithm~\ref{alg:alg_red}summarizes both the forward and backward procedures of the reversible fusion module, including how intermediate states are recomputed to achieve memory efficiency. , the reversible updates deterministically reconstruct earlier states $(f_t, f_{t-1})$ from $(f_{t+1}, f_t)$. During backpropagation, we recompute these intermediate states instead of storing activations from the forward pass, which substantially reduces memory usage, which is a key benefit under limited GPU budgets. 

\renewcommand{\algorithmicrequire}{\textbf{Input:}}
\renewcommand{\algorithmicensure}{\textbf{Output:}}
\begin{algorithm}[H]  
	\caption{RED: The Function BLOCK REVERSE} 
	\label{alg:alg_red} 
	\begin{algorithmic}[1]
        % \STATE it=ft, vt=ft+1
		\REQUIRE: $(f_t,f_{t+1})$, $(\bar{f}_t, \bar{f}_{t+1})$
		\ENSURE: $f_{t-1}$, and $\bar{\omega}_{\mathcal{F}}$ 
        \STATE $f _ {t-1} \leftarrow f _ {t+1} - \mathcal{F} _ t(f_t)$ \COMMENT{Get the old state by subtracting a function.}
		\STATE $\bar{f} _ {t-1} \leftarrow \bar{f} _ {t+1} + \left( \frac{\partial \mathcal{F}}{\partial f _ t} \right)^\top \bar{f} _ {t-1}$ \COMMENT{Update gradient for current using backward rule.}
        \STATE $\bar{\omega} _ {\mathcal{F}} \leftarrow \left( \frac{\partial \mathcal{F}}{\partial \omega _ {\mathcal{F}}} \right)^\top f_t$ \COMMENT{Get the gradient of the parameters.}
        \STATE $f _ t \leftarrow f _ {t-1}$  \COMMENT{Copy the old state to current.}
        % \STATE $\bar{f}_{t-1} \leftarrow \bar{f}_t$
        \STATE $f _ {t+1} \leftarrow f _ t$ \COMMENT{Copy current state to next.}
        \STATE return $f _ {t-1}$, $\bar{f} _ {t-1}$, $\bar{\omega} _ {\mathcal{F}}$ \COMMENT{Give back the state and gradients.}
	\end{algorithmic} 
\end{algorithm}

\subsection{More Ablation Studies}

\begin{figure}
\centering
\includegraphics[width=1\linewidth]{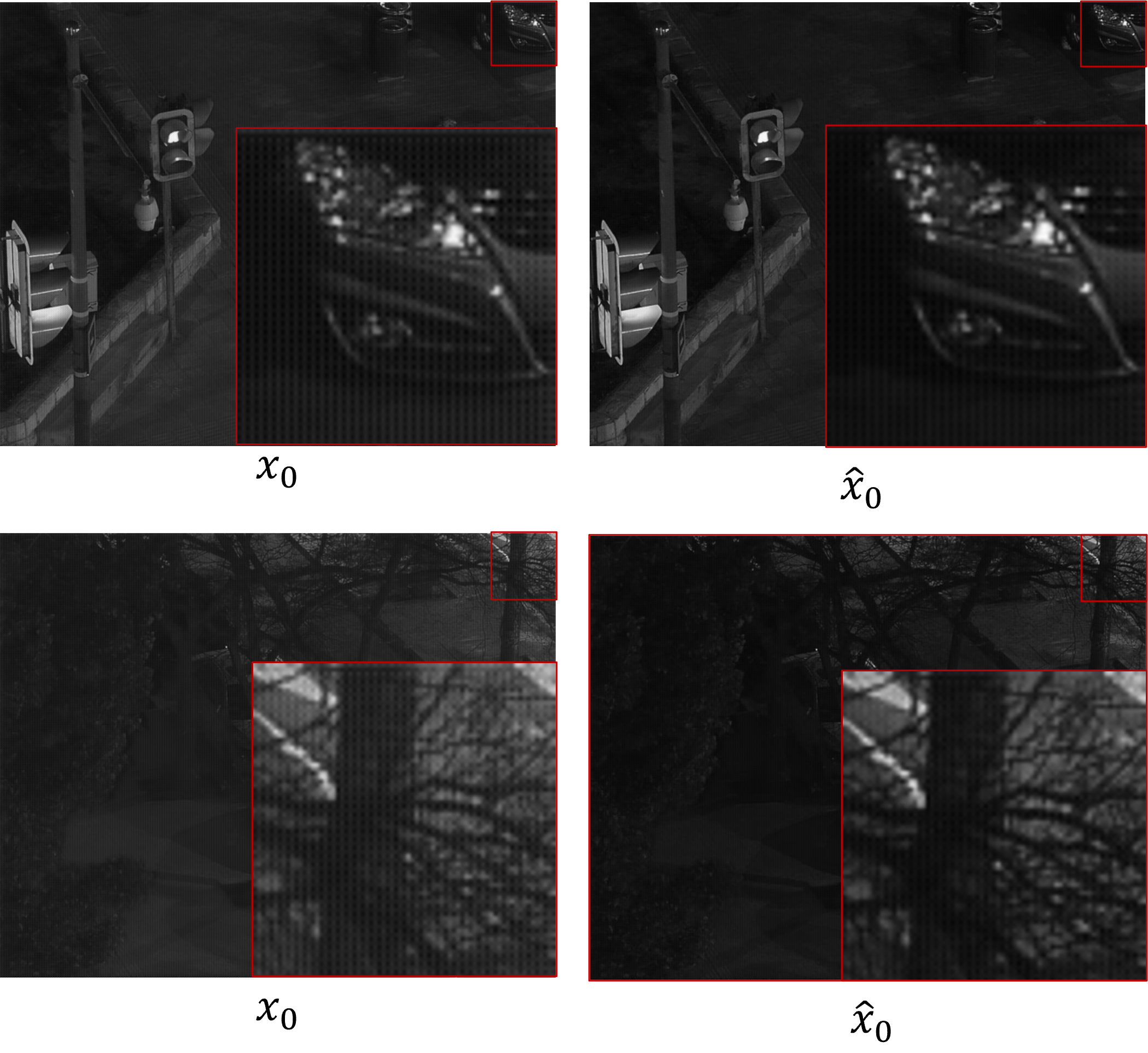}%%%%%%%%%%%%%%%%scale=缩小比例，或者用width=2in
\vspace{-12pt}
\caption{Visualization of image reconstruction results after applying DDPM.}     
\vspace{-12pt}
\label{fig:app_abla_1}
\end{figure}

Fig.~\ref{fig:app_abla_1} shows the results reconstruction after DDPM. In the RED ablation study, we conducted experiments using DDPM to highlight the differences between RED and a simple cascaded model. Furthermore, visualizations demonstrate that even a single-step denoising operation within the diffusion framework can substantially improve the quality of the generated images.

RED employs an iterative fusion strategy, progressively refining the fusion process, where $f_T$ is the refined version of $f_{T-1}$. The final fused image is obtained using $f = w \cdot f_T + (1 - w) \cdot f_{T-1}$, where $w$ is a learnable parameter optimized during training and used during inference, which supplements the information from the previous step.
Table~\ref{tab:app_abla} presents additional ablation results regarding the learnable parameter $w$. Ablation (1) demonstrates that the learnable parameter $w$ can adaptively determine an appropriate fusion weight, effectively combining results of different degrees. Table~\ref{tab:training_viff} shows the values of $w$ across different training epochs and fusion steps, illustrating its interpretability. The $w$ starts at 1 and slightly decreases with the training. Since $f_T$ is much more important than $f_{T-1}$, $w$ places a stronger emphasis on $f_T$. Over time, RED learns to integrate both $f_T$ and $f_{T-1}$, with $f_T$ continuing to dominate the final fusion process. $f_T$ is the refined version of $f_{T-1}$ and shows higher performance compared to $f_{T-1}$. The fused output, integrating both images with the learned weight, further enhances performance.

Furthermore, we replaced the pixel and pixel shuffle operations (used for down- and up-sampling, respectively) with convolution and transposed convolution layers, as commonly done in VAE. However, convolution-based sampling is more difficult to train, prone to collapse, and introduces additional learnable parameters. Ablation (2) shows that RED performs better with pixel shuffle-based sampling, indicating that this approach is more stable and effective for fusion tasks.

\begin{table*}
\centering
% \vspace{-20pt}
\tabcolsep=4pt
% \footnotesize
\caption{Evolution of fusion weight $w$ and VIFF scores across training epochs.}
\fontsize{9}{10}\selectfont\setlength{\tabcolsep}{1.2mm}
\vspace{-5pt}
\scalebox{1}{
\begin{tabular}{lcccccc}
\toprule
Training epoch & $0$ (initial) & $2$ & $4$ & $6$ & $8$ & $10$ \\
\midrule
$w$              & $1.0000$ & $0.9662$ & $0.9617$ & $0.9595$ & $0.9579$ & $0.9567$ \\
$f_{T-1}$ (VIFF) & $0.4567$ & $0.4937$ & $0.4937$ & $0.4686$ & $0.4475$ & $0.4375$ \\
$f_{T}$ (VIFF)   & $0.8650$ & $0.9062$ & $0.9277$ & $0.9336$ & $0.9329$ & $0.9298$ \\
$f$ (VIFF)       & $0.8661$ & $0.9078$ & $0.9293$ & $0.9360$ & $0.9350$ & $0.9320$ \\
\bottomrule
\end{tabular}
\label{tab:training_viff}
}
\vspace{-3pt}
% \end{wraptable}
\end{table*}

\begin{table*}[ht]
\centering
\caption{Ablation on the number of diffusion steps $T$.}
\fontsize{9}{10}\selectfont\setlength{\tabcolsep}{1.2mm}
\vspace{-5pt}
\scalebox{1}{
\begin{tabular}{lcccccccc}
\toprule
Setting & EI & AG & SF & QABF & VIFF & Mem.(GB) & Inf.(s) & Tr.(h) \\
\midrule
T=2 (2$\times$)   & 14.161 & 5.524 & 19.014 & 0.780 & 1.054 & 7.14 & 27.0 & 40 \\
T=2               & 14.742 & 5.908 & 19.293 & 0.744 & 0.931 & 7.14 & \textbf{1.3} & 40 \\
T=4               & 14.832 & 5.945 & 19.457 & 0.746 & 0.942 & 7.15 & 2.7 & 80 \\
T=6               & 14.798 & 5.939 & 19.455 & 0.747 & 0.941 & 7.15 & 5.9 & 120 \\
T=8               & \textbf{14.846} & \textbf{5.958} & \textbf{19.540} & \textbf{0.748} & \textbf{0.944} & 7.14 & 8.0 & 160 \\
\bottomrule
\end{tabular}
}
\label{tab:step}
\end{table*}

\begin{table*}[ht]
\centering
\caption{Quantitative comparison of diffusion-based methods.}
\fontsize{9}{10}\selectfont\setlength{\tabcolsep}{1.2mm}
\vspace{-5pt}
\scalebox{1}{
\begin{tabular}{c|ccccccc}
\toprule
Method & EI & AG & SF & QABF & VIFF & Mem.(GB) & Tr.(h) \\
\midrule
CCF         & 7.10  & 5.02 & 16.83 & 0.42 & 0.50 & --   & --   \\
Text-DiFuse & 12.56 & 4.85 & 15.53 & 0.40 & 0.52 & 22.8 & 381  \\
LFDT-Fusion & 14.26 & 5.71 & 18.96 & 0.72 & 0.89 & 11.8 & 44   \\
RED (ours)  & \textbf{14.74} & \textbf{5.91} & \textbf{19.29} & \textbf{0.74} & \textbf{0.93} & \textbf{7.14} & \textbf{40} \\
\bottomrule
\end{tabular}
}
\label{tab:fusion_comparison_diffusion}
\end{table*}

\begin{table}
\centering
% \vspace{-20pt}
\tabcolsep=4pt
% \footnotesize
\caption{Ablation studies on the LLVIP dataset. \textbf{Bold} indicates the best performance. (1) denotes the variant without the learnable parameter $w$; (2) replaces the up- and down-sampling operations with convolution and transposed convolution; (3) corresponds to the RED model.}
\fontsize{9}{10}\selectfont\setlength{\tabcolsep}{2.6mm}
\vspace{-5pt}
\scalebox{1}{
\begin{tabular}{c|ccccccc}
\toprule
 & EI   & AG  & PSNR &  $Q^{AB/F}$  & VIFF \\
\midrule  
(1) & $14.58$ & $5.85$ & $19.20$ & $0.74$ & $0.93$ \\
(2) & $14.52$ & $5.82$ & $19.03 $& $0.74$ & $0.93$ \\
(3) & $\mathbf{14.74}$ & $\mathbf{5.91}$ & $\mathbf{19.29}$ & $\mathbf{0.74}$ & $\mathbf{0.93}$ \\	
\bottomrule
\end{tabular}
}
\vspace{-3pt}
\label{tab:app_abla}
% \end{wraptable}
\end{table}

\begin{figure*}
\centering
\includegraphics[width=0.95\linewidth]{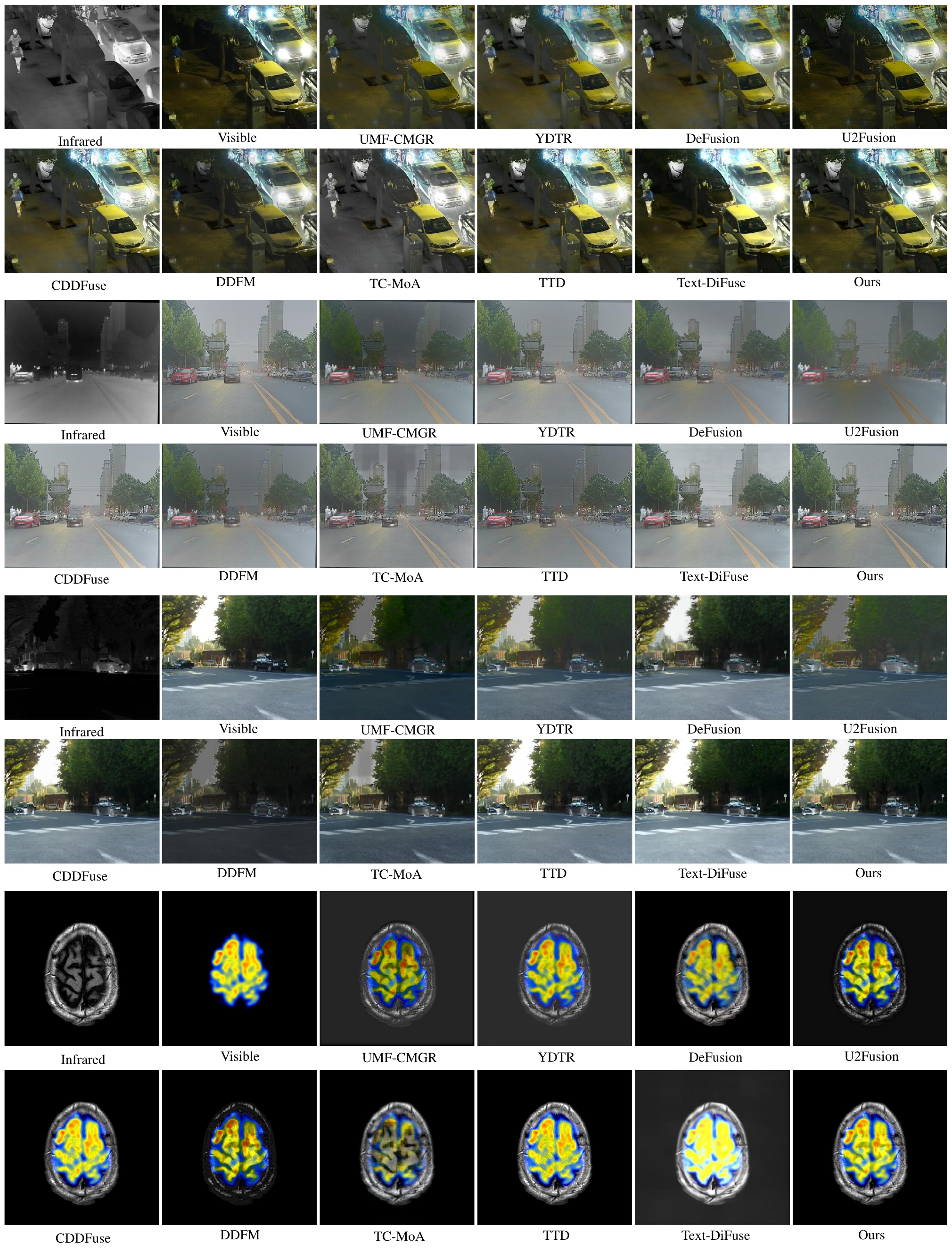}%%%%%%%%%%%%%%%%scale=缩小比例，或者用width=2in
\vspace{-5pt}
\caption{Visualization comparing SOTA methods.}     
\vspace{-12pt}
\label{fig:app_more_comp}
\end{figure*}
\begin{figure*}
\centering
\includegraphics[width=0.9\linewidth]{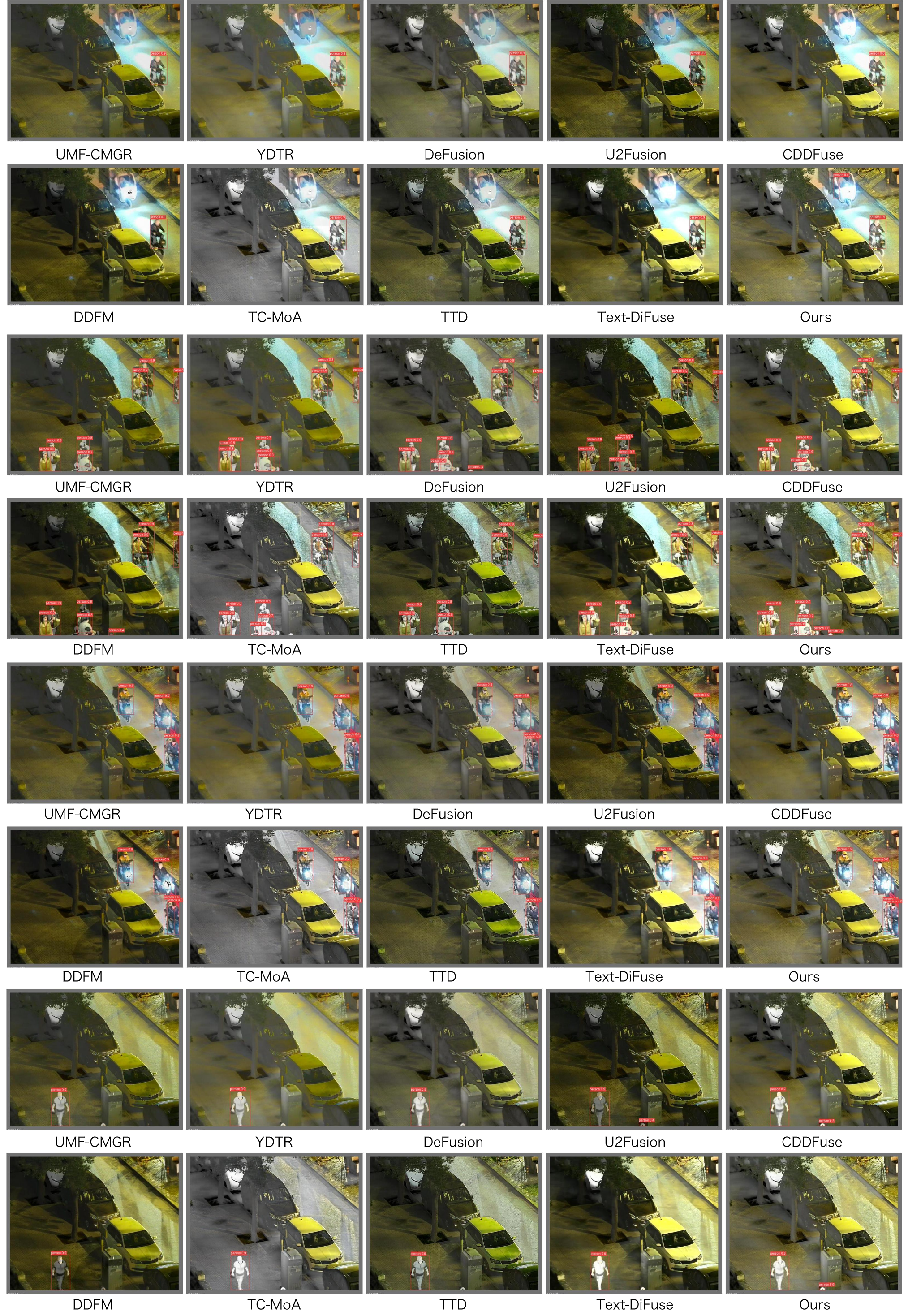}%%%%%%%%%%%%%%%%scale=缩小比例，或者用width=2in
\vspace{-8pt}
\caption{Visualization comparing SOTA methods on the object detection task.}     
\vspace{-12pt}
\label{fig:det4}
\end{figure*}
\begin{figure*}
\centering
\includegraphics[width=1\linewidth]{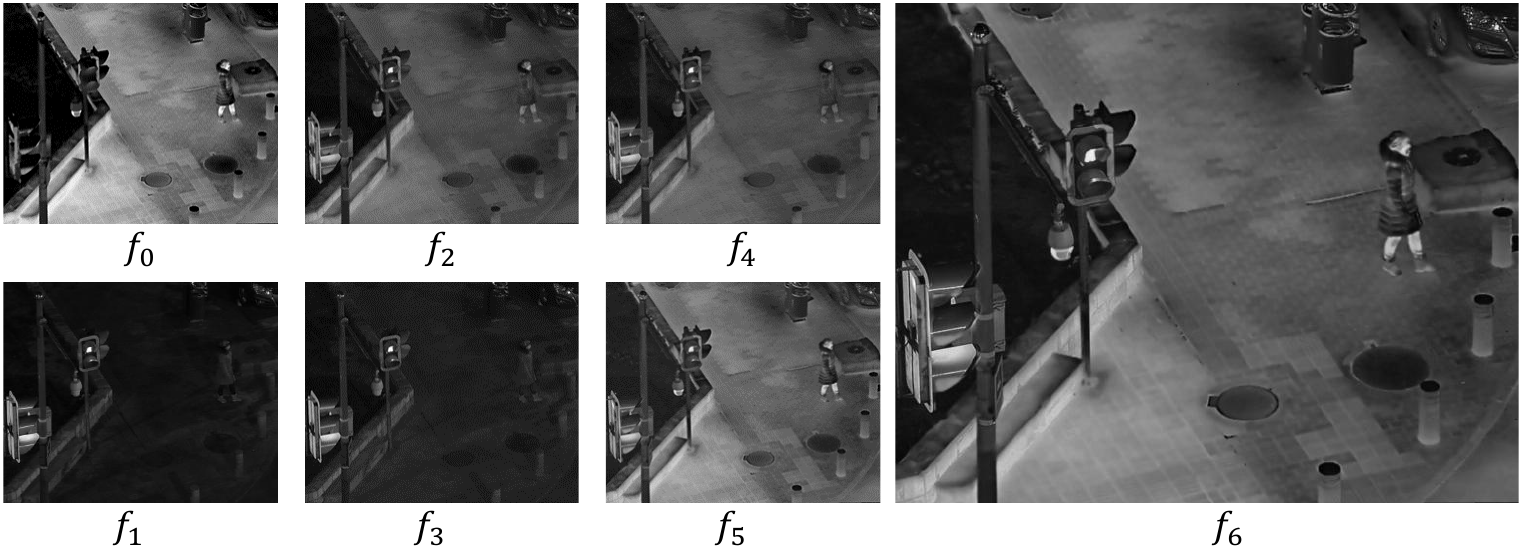}%%%%%%%%%%%%%%%%scale=缩小比例，或者用width=2in
\vspace{-8pt}
\caption{Visualization of Intermediate outputs during the inference at the diffusion step 6.}     
\vspace{-12pt}
\label{fig:vis_intermediate}
\end{figure*}
% 此处放折线图或者表格
\subsection{More Preliminaries}
\textbf{DDPM.}
Denoising Diffusion Probabilistic Models (DDPM) construct a generative framework that has demonstrated remarkable performance in various image synthesis tasks. The model consists of a forward diffusion process that gradually corrupts a clean sample with Gaussian noise, and a corresponding reverse denoising process that iteratively reconstructs the original data. In the forward process, a clean data sample $ x_0 \sim p_\text{data}(x)$ is transformed into a noisy latent variable $ x_T$ over $ T$ steps, discrete time steps according to the following stochastic equation:
\begin{equation}
    x_t = \sqrt{\alpha_t} x_{t-1} + \sqrt{1-\alpha_t} \epsilon_t, \quad \epsilon_t \sim \mathcal{N}(0, I)
\end{equation}

with $ \alpha_t := 1-\beta_t $, and $ \{\beta_t\}_{t=1}^T $ is a predefined noise schedule. This formulation admits a closed-form solution for arbitrary time steps:
\begin{equation}
x_{t-1}=\frac{1}{\sqrt{\bar{\alpha}_t}}(x_t-\frac{\beta_t}{\sqrt{1-\bar{\alpha}_t}}\tilde{\epsilon}), \quad \bar{\alpha}_t = \prod_{i=1}^t \alpha_i,
\end{equation}
 where $\tilde{\epsilon}=\epsilon_\theta(x_t, t)$ estimated noise with a neural network. $\sigma_t$ is the standard deviation within the sampling process.
The reverse process learns a parameterized conditional distribution $p_\theta(x_{t-1}|x_t) $ that approximates the true posterior of the denoising process. To achieve this, the model is trained to predict the noise term $\epsilon$ using a neural network $\epsilon_\theta(x_t, t) $ by minimizing the following objective:
\begin{equation}
 \min_\theta \mathbb{E}_{x_0, \epsilon, t} \left[ \| \epsilon - \epsilon_\theta(x_t, t) \|^2 \right]
\end{equation}
Once trained, the model can generate new samples by starting from a noise vector $ x_T \sim \mathcal{N}(0, I)$ and applying the learned reverse process iteratively.  

\subsection{More Implementation Details}
We use Stable Diffusion v1.5 (SD 1.5) as the backbone, with most of the blocks pruned to reduce computational complexity. The RED model is initialized with the pretrained parameters from SD v1.5 and fine-tuned with $T = 2$ diffusion steps, a batch size of 20, and 10 training iterations on a single NVIDIA A6000 GPU (38 GB memory). The $w$ is initialized by 1.

\subsection{More Details about Datasets}
\label{sec:data_detail}
Below, we provide a detailed description of each dataset:

\textbf{LLVIP}~\cite{jia2021llvip} is a visible-infrared paired dataset for low-light vision tasks. It contains $15,488$ pairs, most of which were captured in very dark scenes. All images are strictly aligned in both time and space. 

 \textbf{MSRS}~\cite{tang2022piafusion} is a multi-spectral dataset for infrared and visible image fusion, containing $1,444$ pairs of high-quality, aligned infrared and visible images. It is derived from the MFNet dataset, with $715$ daytime image pairs and $729$ nighttime image pairs collected after removing misaligned pairs. 

 \textbf{M3FD}~\cite{liu2022target} is constructed using a synchronized acquisition system equipped with a binocular optical camera and a binocular infrared sensor. It includes $4,200$ image pairs for training and $300$ independent scenes for testing. All visible images are calibrated using the system's internal parameters, while infrared images are artificially distorted via a homography matrix to simulate real-world challenges. The dataset is particularly valuable for evaluating the generalization capability of fusion methods across diverse scenarios.

\subsection{More Visualization of Results}
Additional visual comparisons are presented in Fig.~\ref{fig:app_more_comp}. RED preserves texture details more effectively than Text-DiFuse, another diffusion-based method. This advantage stems from RED’s design, which incorporates direct supervision into the diffusion process. Compared with other methods, RED consistently demonstrates strong performance in preserving information across both high-light and low-light regions in all three infrared and visible-light datasets. This capability is enabled by RED’s iterative optimization strategy, which progressively refines the fusion process. By leveraging the complementary strengths of both modalities, RED generates the most balanced and visually coherent fused images.

\subsection{More Visualization of Downstream}
We show some visual samples of object detection in the LLVIP dataset in Fig.~\ref{fig:det4}. RED also performs effectively in downstream tasks such as object detection. It successfully detects objects near occluded or cluttered edges—areas often missed by other methods, and accurately identifies dense, complex crowd regions. These visual results demonstrate that RED significantly improves detection accuracy in challenging scenarios.

\subsection{Analysis of Intermediate Outputs}
$f_T$ is a refined state of $f_{T-1}$ and empirically achieves higher performance than $f_{T-1}$. The final fused image $\hat{y}$ integrates information from both states via learned fusion weights $w$, further improving perceptual quality and cross-modal consistency.
Figure~\ref{fig:vis_intermediate} visualizes the intermediate states $f_{T-1}$ and $f_T$. As diffusion progresses, information from one modality is progressively injected into the other, yielding a coherent representation; the progressive fusion process gradually integrates complementary modality-specific details into a single, comprehensive image.

\subsection{Comparison With Diffusion-based Methods}
As shown in Table~\ref{tab:fusion_comparison_diffusion}, we report a quantitative comparison against recent latent-diffusion–based fusion methods, including CCF (training-free), Text-Diffuse (built on DDFM), and LFDT-Fusion (LDM-based). RED consistently outperforms these latent-diffusion baselines.

We attribute RED’s gains to its non-Markov reverse process and reversible fusion framework, which together improve fidelity and information preservation in the fused images. In addition, RED offers notable memory efficiency during training: the reversible mechanism allows us to recompute intermediate states in backpropagation instead of storing them, reducing GPU memory usage at a given batch size and improving overall training efficiency.
In terms of wall-clock performance, RED’s training speed is comparable to LFDT-Fusion: our end-to-end formulation converges quickly, whereas LFDT-Fusion incurs extra steps due to its diffusion procedure, leading to longer training times. This time–memory trade-off stems from our reversible design, which intentionally spends additional computation to save memory.

In summary, RED achieves superior accuracy while using fewer computational resources than current diffusion-based fusion methods.

\subsection{More exploration}

As detailed in the main text and illustrated in Fig. 7 (Page 9), we further analyzed model performance under increased diffusion steps. The results in Table~\ref{tab:step} indicate a general performance improvement as the number of diffusion steps increases. Memory consumption does not grow significantly with additional steps; however, both inference and training time increase proportionally due to the reversible fusion design, where more steps naturally require additional computation.

In addition, our model supports images of various resolutions. As shown in Table~\ref{tab:step}, when testing with higher-resolution inputs (2×), we observe slight decreases in EI, AG, and SF, while $Q^{AB/F}$ and VIFF improve. The improvement in $Q^{AB/F}$ and VIFF can be attributed to the preservation of more fine-grained modality-specific details at higher resolutions, whereas the minor reductions in the other metrics may result from gradient dispersion introduced by increased spatial resolution. Although memory usage remains nearly unchanged, inference time increases substantially because the attention mechanism in the SD network has computational complexity that scales quadratically with image resolution.

% \begin{table*}[ht]
% \centering
% \caption{Ablation on the number of diffusion steps $T$.}
% \fontsize{9}{10}\selectfont\setlength{\tabcolsep}{1.2mm}
% \vspace{-5pt}
% \scalebox{1}{
% \begin{tabular}{lcccccccc}
% \toprule
% Setting & EI & AG & SF & QABF & VIFF & Mem.(GB) & Inf.(s) & Tr.(h) \\
% \midrule
% T=2 (2$\times$)   & 14.161 & 5.524 & 19.014 & 0.780 & 1.054 & 7.14 & 27.0 & 40 \\
% \textbf{0.944} & 7.14 & 8.0 & 160 \\
% \bottomrule
% \end{tabular}
% }
% \label{tab:resolution}
% \end{table*}

\subsection{Limitations and Broader Impacts}
\label{sec:limitation}
Although the proposed RED model achieves superior performance compared to existing methods and demonstrates strong generalization across diverse scenarios and tasks, several potential limitations remain. We introduce a reversible fusion paradigm that significantly reduces memory usage, enabling end-to-end training of diffusion models without relying on Markov chains and thereby avoiding error accumulation that often leads to detail distortion. However, this memory-saving approach adopts a time-for-space trade-off, which results in slower training times. Additionally, the choice of the number of diffusion steps is currently empirical. While we observe that increasing the number of steps generally improves performance, an optimal step count cannot be precisely determined and may vary across tasks. From a broader societal perspective, the inability to guarantee an optimal trade-off between computational cost and performance could pose risks in high-stakes applications such as medical imaging and autonomous driving, where reliability and efficiency are critical.

{
    \small
    \bibliographystyle{ieeenat_fullname}
    \bibliography{main}
}

% WARNING: do not forget to delete the supplementary pages from your submission 
% \input{sec/X_suppl}

{
    \small
    \bibliographystyle{ieeenat_fullname}
    \bibliography{main}
}

% WARNING: do not forget to delete the supplementary pages from your submission 
% \input{sec/X_suppl}

\end{document}